\documentclass[journal]{IEEEtran}
\usepackage{cite}
\ifCLASSINFOpdf
  \usepackage[pdftex]{graphicx}
\else
\fi
\usepackage{algorithmic}
\ifCLASSOPTIONcompsoc
  \usepackage[caption=false,font=normalsize,labelfont=sf,textfont=sf]{subfig}
\else
  \usepackage[caption=false,font=footnotesize]{subfig}
\fi

\usepackage{amssymb}
\usepackage{amsmath}
\usepackage{amsfonts}

\usepackage{url}
\usepackage{multirow}
\usepackage{bm}

\usepackage{enumitem}

\usepackage{hyperref}

\usepackage{color, colortbl}
\usepackage{xcolor}
\definecolor{lightgray}{gray}{0.1}
\definecolor{orange}{rgb}{1,0.5,0}

\usepackage{fancyhdr}
\begin{document}

\title{On Social Interactions of Merging Behaviors at Highway On-Ramps in Congested Traffic}

\author{Huanjie~Wang,
        Wenshuo~Wang,~\IEEEmembership{Member,~IEEE,}
        Shihua~Yuan,
        Xueyuan~Li,
        and~Lijun~Sun % <-this % stops a space
\thanks{This work is supported by the National Natural Science Foundation of China (Grant No. U1864210), and the China Scholarship Council (No. 201906030060). (\textit{Corresponding Authors: Wenshuo Wang and Shihua Yuan})}

\thanks{H. Wang is with the School of Mechanical Engineering, Beijing Institute of Technology, Beijing 100081, China (e-mail: wanghj\_815@163.com).

W. Wang is with the University of California at Berkeley, Berkeley, CA 94720 USA (e-mail: wwsbit@gmail.com).}% <-this % stops a space
\thanks{S. Yuan and X. Li are with the School of Mechanical Engineering, Beijing Institute of Technology, Beijing 100081, China (e-mail: yuanshihua@bit.edu.cn).}% <-this % stops a space
\thanks{L. Sun is with the Department of Civil Engineering and Applied Mechanics, McGill University, Montreal, QC H3A 0C3, Canada (e-mail: lijun.sun@mcgill.ca).}%

\thanks{Digital Object Identifier 10.1109/TITS.2021.3102407}
}

%\markboth{IEEE Transactions on Intelligent Transportation Systems,~Vol.~xxx, No.~xxx, xxx~2020}{Shell \MakeLowercase{\textit{et al.}}: Bare Demo of IEEEtran.cls for IEEE Journals}

% after the title page when using the two side option.

\maketitle

\thispagestyle{fancy}


\begin{abstract}
Merging at highway on-ramps while interacting with other human-driven vehicles is challenging for autonomous vehicles (AVs). An efficient route to this challenge requires exploring and exploiting knowledge of the interaction process from demonstrations by humans. However, it is unclear what information (or environmental states) is utilized by the human driver to guide their behavior throughout the whole merging process. This paper provides quantitative analysis and evaluation of the merging behavior at highway on-ramps with congested traffic in a volume of time and space. Two types of social interaction scenarios are considered based on the social preferences of surrounding vehicles: \textit{courteous} and \textit{rude}. The significant levels of environmental states for characterizing the interactive merging process are empirically analyzed based on the real-world INTERACTION dataset. Experimental results reveal two fundamental mechanisms in the merging process: 1) Human drivers select different states to make sequential decisions at different moments of task execution, and 2) the social preference of surrounding vehicles can impact variable selection for making decisions. It implies that efficient decision-making design should filter out irrelevant information while considering social preference to achieve comparable human-level performance. These essential findings shed light on developing new decision-making approaches for AVs.
\end{abstract}

% Note that keywords are not normally used for peer review papers.

% \fi
\IEEEpeerreviewmaketitle

\section{Introduction} \label{sec:introduction}

\begin{figure}[t]
\centering
\includegraphics[width=\linewidth]{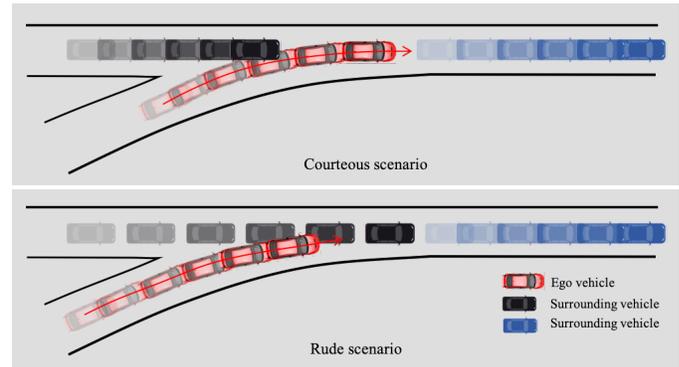}
\caption{Illustration of the two typical highway on-ramp merge scenarios with different social preferences of the surrounding vehicles: courteous (top) and rude (bottom). Top: The black vehicle behaves courteously and allows the ego vehicle to cut in. Bottom: The black vehicle behaves rudely and forces the ego vehicle to yield.}
\label{fig:scenario}
\end{figure}

\IEEEPARstart{M}{erging} at highway on-ramps with congested traffic is a daily routine but a challenging task in the real world. Understanding the seemly mundane merging processes demonstrated daily by human drivers is critical for autonomous vehicles (AVs) that can safely and efficiently interact with humans around them \cite{zgonnikov2020should}. In mixed traffic, AVs must respond to contextual changes effectively. Inefficient collaboration with its surrounding humans can cause typical traffic issues such as oscillations, congestion, and speed breakdown \cite{marczak2013merging, sun2014modeling}. According to the recent report by the National Highway Traffic Safety Administration (NHTSA) \cite{national2018summary}, nearly 30,000 highway merging collisions occur each year in the USA. The desired merging execution should guarantee traffic safety while avoiding congestion. Fig. \ref{fig:scenario} shows a typical highway on-ramp merge scenario. The merging vehicle (denoted as the ego vehicle) runs in a merge lane and plans to merge into the mainstream traffic flow on the highway while interacting with the surrounding cars. The ego vehicle needs to make a sequential decision according to their \textit{situation-awareness}\footnote[1]{Situation-awareness is formally defined as a person's \textit{perception} of the elements of the environment within a volume of time and space, the \textit{comprehension} of their meaning and the \textit{projection} (also known as prediction) of their status in the near future \cite{endsley1995measurement}}, which is essentially related to the augmented perceptual information of the environmental states. The augmented perceptual information consists of direct perceptional information about the environment and indirect inferred causes, and both are intrinsically dynamic and stochastic. The most used perceptual information for decision-making and task execution of merging at highway on-ramps includes the state of the ego vehicle and surrounding vehicles and their variants, such as their relative gaps and time-headway. However, the above-selected variables might vary over time and space and be influenced by other human drivers' social preferences, for example, competitive and prosocial.

% scenario

On the one hand, the social preference of the surrounding vehicles might impact the ego vehicle's decision. Considerable efforts of decision-making algorithms have been made to execute the merging task at highway on-ramps. The surrounding human drivers make the merging task more challenging due to the variety of social preferences \cite{schwarting2019social, koechlin2020human}. For instance, in some scenarios (the top case in Fig. \ref{fig:scenario}), the surrounding vehicle behaves courteously by slowing down and leaving an acceptable gap ahead to allow the ego vehicle to cut in. In some other scenarios (the bottom case in Fig. \ref{fig:scenario}), the surrounding vehicle acts rudely by narrowing its gap ahead, thus forcing the ego vehicle to select the next gap for cutting in. Besides, studies of human social decision-making under controllable laboratory settings demonstrate that humans make sequential decisions and execute specific tasks based on different preferences \cite{seo2012neural}. However, it is unclear whether the mechanisms implicated in simple social decisions in the laboratory are paramount in complex driving scenarios, such as merging at highway on-ramps.

On the other hand, the informational needs of human drivers for achieving the task vary over time and space \cite{leong2017dynamic, niv2019learning, radulescu2021human, weng2018time}. In real-life situations, generic laboratory evidence reveals that some environmental features are not readily observed but are critical for guiding appropriate behavior\cite{langdon2019uncovering}. In contrast, others are salient but irrelevant for task performance\cite{yang2018scene}. Humans do not utilize all of the sensory information (or the whole state of the environment), but only a few aspects of it, to make decisions \cite{wilson2012inferring}. 
For example, a human driver would use some critical states of the environment to decide which gap is acceptable in merging behavior and then dilute the significance of some of them over time and space to execute other subtasks. Therefore, an empirical investigation into the relevance (i.e., significant variables and saliency) of the merging scenarios at highway on-ramps is much needed.

The above analysis signifies that in a highway merging scenario with congested traffic, the ego vehicle's decision-making relies on the social preference of the surrounding vehicles. However, there are still unclear mechanisms reflected by three fundamental questions: 

\begin{itemize}
\item What are the relevances (or the states of the environment) to guide human drivers to make a decision?
\item How are the relevances (or the states of the environment) changing over time and space?
\item How do the social preferences of the surrounding humans make these relevances different?
\end{itemize}

Based on the above questions, this paper conducts a comprehensive analysis using a real-world dataset. This paper aims to bridge the gap between the merging driver's decision-making and the surrounding human drivers' social preferences over time and space. To this end, we extract the merging scenarios from the real-world dataset and classify them into two groups according to the surrounding vehicles' social preferences. We then quantitatively analyze and evaluate how the ego vehicle utilizes the environmental states to make decisions over the merging process. Finally, we provide a comparison to reveal that social preference can impact on the ego vehicle's decision-making.

The remainder of this paper is organized as follows. Section II reviews related works on merging behaviors, variable selection, and social behavior. Section III defines the merging scenarios. Section IV discusses real-world datasets, data preprocessing, and methods. Section V analyzes the experimental results and provides further discussion. Finally, Section VI gives the conclusion.

\section{Related Works}
In this section, we first analyze the development of merging behavior and variable selection. Then, we review the related research on social behavior. Finally, we make a summary based on the above two aspects.

\subsection{Merging Behavior \& Variable Selection}

In general, merging behavior can be distinguished as discretionary (changing the driving conditions) and mandatory (reaching the target lane/destination). Highway on-ramp merge is a typical mandatory merging task with social interactions. Much research on highway merging has been conducted. At first, the gap acceptance theory is the most commonly used one by assuming that the merging vehicle can merge into the target lane if the gap between the assumptive lead and lag vehicles is acceptable \cite{ahmed1996models, lee2006modeling, toledo2009estimation}. However, this is often not the case in practice. The vehicle will still initiate merging behavior even if the selected gap is less than the critical value. This situation usually occurs at highway on-ramps in congested traffic. The merging vehicles are supposed to merge into the main road as quickly as possible, so their tentative merge intention will still occur even if the selected gap does not meet the distance requirements \cite{sun2014driver, zheng2014recent, tang2018deviation}. Researchers in \cite{kita1993effects} developed a binary logistic model to describe the probability of merging decisions with the gap, vehicle speed, and remaining distance to the end of the ramp, and acceleration \cite{weng2011modeling, marczak2013merging,fatema2013probabilistic}.

To consider the heterogeneity among drivers, Weng, \textit{et al}. \cite{weng2015depth} developed a mixed probabilistic merging model by introducing two additional surrogate safety measures: time to collision (TTC) and deceleration rate to avoid crash (DRAC). After that, they applied a finite mixture of the logistic regression model to analyze the heterogeneity for different drivers based on the gap, (relative) speed/distance, distance to the start of the ramp, and whether a lead vehicle exists in the merge lane \cite{li2018application}. 

The time-varying effects of different variables are also considered to describe dynamic merging behavior. Researchers in \cite{mccall2007lane} selected the lane position information, vehicle parameters, and head motion to infer driver intent. The Cellular Automata models incorporating dynamic behavior were also developed in \cite{meng2011improved}. From the perspective of safety and collision avoidance, Weng, \textit{et al}. developed a time-varying mixed logit regression model to describe the merging process and analyzed the time-varying effects of variables on merging behavior, including vehicle speed, vehicle type, and crash probability and severity \cite{weng2018time}. They claimed that vehicle crash probability and severity were the contributory factors instead of vehicle speeds and gap sizes. 

With the fact that the surrounding vehicle also needs to respond to the merging behavior, researchers in \cite{arbis2019game} applied game theory to make the interaction more understandable. In recent years, data-driven methods have drawn more attention. For example, classification and regression trees (CART), Bayesian network (BN), and fuzzy logic models were used in developing merging decisions based on variables including relative speed, lead/lag gap, and remaining distance \cite{meng2012classification, hou2013modeling, tang2018lane}.

\subsection{Social Behavior}
Humans will consider social interactions rather than their own individual goals when making decisions \cite{seo2012neural}. It is an essential capability for autonomous vehicles to identify social behavior and thus make interpretable decisions accurately. Usually, the environmental states in the real world are partially observable and dynamic. They can be formulated via a time-dependent partially observable Markov decision process (POMDP) to improve naturalness and social propriety \cite{broz2008planning}. Leveraging social conventions into the optimization constraints could improve path planning and navigation performance \cite{kirby2010social}. Wei, \textit{et al}. \cite{wei2013autonomous} proposed a Bayesian-based social behavior framework to predict other agents' intentions, thus enabling more sociable decisions of the autonomous system. Sun, \textit{et al}. \cite{sun2018courteous} introduced courteous planning to reduce the inconvenience of human drivers and benefit both sides. Ren, \textit{et al}.\cite{ren2019shall} proposed a model predictive control method to tackle the two-player game, allowing autonomous vehicles to learn more social behaviors based on social grace. By considering both rational and irrational social behaviors, Hu, \textit{et al}. \cite{hu2019generic} presented a prediction framework to estimate the continuous trajectories of surrounding vehicles. By defining the optimal control problem and formulating the appropriate algorithm, Speidel, \textit{et al}. \cite{speidel2019towards} proposed a planning framework to avoid being too aggressive. Schwarting, \textit{et al}. \cite{schwarting2019social} borrowed the Social Value Orientation (SVO) from the field of social psychology to quantify the degree of selfishness or altruism, which provides the basis for solving dynamic games in a socially acceptable way.

\subsection{Summary}

In terms of decision making and variable selection, almost all the existing research on merging behavior only concerns static analysis. Still, it merely considers the dynamic dominant states of the environment over the merging process. However, it is not clear whether the effects of dominant states may change over time and space, which is also one of the fundamental problems. Research in \cite{weng2018time} analyzed the time-varying impacts of different variables on merging behavior. However, their significant analysis was based on a univariate statistical technique, i.e., analyzing a single variable while fixing other variables. This analysis could cause biased conclusions since multiple variables can influence the merging behavior simultaneously in reality. Besides, they neglected the influence of human drivers with various social preferences on variable selection. More specifically, they did not consider the interactions between the merging vehicle and the \textit{rude} social preferences of the surrounding vehicles. 

For social interaction, individual drivers usually have various social preferences, reflected by differences in reactions and motivations. Therefore, at different moments of the merging process, AVs need to know which variables are necessary to make appropriate decisions. Many studies have made efforts to make AVs more prosocial. However, none of them consider the near-collision or adversarial interaction scenarios in which both human drivers and AVs must be capable of tackling in real traffic.

In summary, the three fundamental questions asked in the introduction remain open in existing research. Answering these three questions can improve the AV's decision performance in highway merging scenarios and ensure better social interactions.

\subsection{Contributions}

Following the review summary and the three fundamental questions proposed in Section I, the main contributions of this paper are threefold:

\begin{itemize}

\item Quantitatively analyzing the merging process of highway on-ramps in congested traffic based on the INTERACTION dataset.

\item Quantitatively analyzing the saliency of different variables that guide drivers to make decisions in different merging scenarios with rude and courteous social preferences of surrounding vehicles. 

\item Providing a practical basis for selecting significant variables for researchers when designing decision-making algorithms for autonomous vehicles merging at highway on-ramps. 

\end{itemize}

\section{Interaction Procedure of Social Merging Behavior}

\begin{figure}[t]
\centering
\subfloat[Courteous scenario]{\label{level1.sub.1} \includegraphics[width=0.98\linewidth]{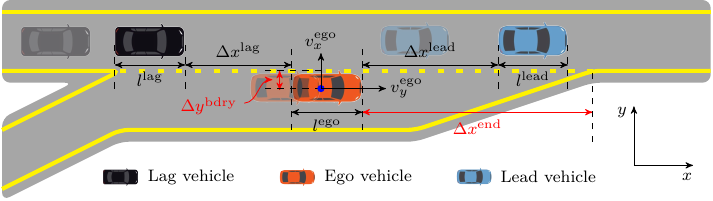}} \\
\subfloat[Rude scenario]{\label{level1.sub.2} \includegraphics[width=0.98\linewidth]{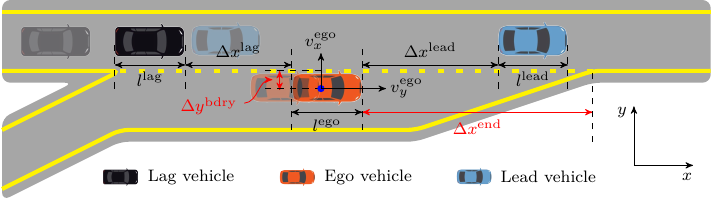}}
\caption{Illustration of the two typical highway on-ramp merge scenarios wherein two surrounding vehicles (lead vehicle and lag vehicle) interact with the ego vehicle. The transparent vehicle represents the initial position of vehicles at the start moment of $t_s$.}
\label{fig:Diagram of the typical highway on-ramp merge scenario}
\end{figure}

In this section, we first specify the two merging scenarios at the highway on-ramp (Fig. \ref{fig:Diagram of the typical highway on-ramp merge scenario}). In this scenario, the ego vehicle travels in the merge lane and intends to merge into the highway on which two assumptive surrounding vehicles move forward. We then defined three critical moments of the merging process to facilitate analysis and finally discussed variable selection.

% Diagram of the typical highway on-ramp merge scenario

\subsection{Definitions of Surrounding Vehicles} \label{subsec: sur_vehicles}

We mainly considered two surrounding vehicles that interact closely with the merging vehicle. To make the concerned scenarios clear, we name these two surrounding vehicles based on their final position to the merging vehicle when the merge task ends.

\begin{itemize}
\item \textit{Lead vehicle}:
When completing the merge task, the nearest vehicle in front of the merging vehicle is denoted as the lead vehicle. During the merging process, the lead vehicle may be located at the left-behind area of the merging vehicle at the beginning, for example, the blue car in the scenario of Fig. \ref{fig:Diagram of the typical highway on-ramp merge scenario}(b). 

\item \textit{Lag vehicle}:
When completing the merge task, the nearest vehicle behind the merging vehicle is denoted as the lag vehicle, for example, the black car in the scenario of Fig. \ref{fig:Diagram of the typical highway on-ramp merge scenario}(a) and (b).

\end{itemize}

\subsection{Critical Moments} \label{subsec:cri_moments}

\begin{figure}[t]
\centering
\includegraphics[width = \linewidth]{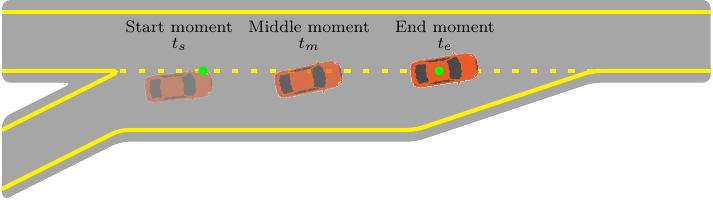}
\caption{An illustration of the three critical moments ($t_s$, $t_m$, and $t_e$) of the merging process at highway on-ramps.}
\label{fig:three_moments}
\end{figure}

Vehicles usually merge into the traffic flow on the highway from on-ramps on the right side because the vehicle speed in the rightmost lane is usually the slowest in countries with right-hand traffic. Hence, this paper mainly focuses on the left merge behavior at highway on-ramp scenarios. In order to study the dynamic changes over the whole merging process, three critical moments (as shown in Fig. \ref{fig:three_moments}) are defined: 

\begin{itemize}
\item \textit{Start moment}:
The start moment $t_s$ refers to when the left front wheel of the ego vehicle crosses the boundary between the merge lane and the rightmost highway lane. 
\item \textit{End moment}:
The end moment $t_e$ refers to when the center of the ego vehicle lies on the boundary between the merge lane and the rightmost highway lane.
\item \textit{Middle moment}:
The middle moment $t_m$ refers to the middle moment between $t_s$ and $t_e$.
\end{itemize}

The interactions between the start and end moments defined as above are particularly strong. More concretely, before the start moment, the ego vehicle's intention is usually vague for the lag vehicle in the target lane. Once the ego vehicle reaches the start moment $t_{s}$, the target lane's lag vehicle can perceive the ego vehicle's merge intention. The lag vehicle can not complete the overtaking behavior in its current lane without colliding with the merging vehicle after the end moment $t_e$. That is to say, the longitudinal relative position relationship between the merging vehicle and its paired lead or lag vehicle will not change. Therefore, the selected information on vehicle ontology and environmental states at the three critical moments allows us to analyze the dynamic interaction process of highway merging.

\subsection{Social Merging Behavior} \label{subsec:social}

In congested highway on-ramp traffic, the gaps between vehicles are small. Therefore, the merging vehicle should actively create a large enough gap by delivering its merge intention to its surrounding vehicles. Although there are many surrounding vehicles in congested traffic, only the lead and lag vehicles in the target lane have strong interaction with the merging vehicle. In other words, social preferences (e.g., courteous and rude) of the lead and lag vehicles will directly influence the merging vehicle's future decisions. As shown in Fig. \ref{fig:scenario}, we mainly focus on two types of social interactions between the merging and assumptive lead/lag vehicles. 

\begin{itemize}

\item \textit{Rude}:
The surrounding vehicle behaves rudely and competes for the right of way with the merging vehicle from the start moment $t_s$. Aggressive or near-collision behavior will be shown in this process because there will be a phenomenon wherein the merging vehicle traveling at a low speed competes with the surrounding vehicle for a while. 

\item \textit{Courteous}:
The surrounding vehicle behaves courteously and gives way to the merging vehicle from the start moment $t_{s}$. Aggressive or near-collision behavior will not occur during this process. 

\end{itemize}

\subsection{Variable Selection} \label{subsec:variable}

Selecting reasonable variables is essential for AVs to make decisions with strong interactions because the environmental information is redundant. Only information directly related to the task is beneficial for AVs to understand the environment and make optimal decisions, while irrelevant information should be removed as noise. For the highway on-ramp merge, the absolute position coordinates of vehicles are unrelated to the task. Instead, relative distance and relative speed should be adopted to capture the relationships between vehicles better. Unlike the absolute position coordinates, the ego vehicle's absolute speed in longitudinal and lateral directions should also be introduced as references to analyze the influence of different traffic conditions on the task. 

Based on previous studies, the surrogate safety measure (SSM) between the ego vehicle and the adjacent surrounding vehicles should also be selected to describe the risk level \cite{oh2010estimation}. Otherwise, it will cause a wrong merge decision in some critical conditions. For example, when the relative speed between the ego vehicle and the lead vehicle on the target lane is relatively large but the relative distance is small, the collision risk is very high. However, the absence of SSM will make it difficult to perceive this high risk. The standard explanatory variables of SSM include the deceleration rate \cite{gettman2003surrogate}, the deceleration rate to avoid the crash (DRAC) \cite{cunto2008calibration}, the time to collision (TTC) \cite{weng2014rear}, and the time headway (THW) \cite{daamen2010empirical}. 

These studies and practices have shown that TTC is more related to risk levels and can reflect the driver's risk preferences. Besides, the use of TTC can also improve the prediction accuracy of the modeling for merge decisions \cite{wang2009study, weng2015modeling}. Therefore, we selected TTC as the influencing variable in this work. Generally speaking, calculating TTC requires that the speed of the vehicle behind is higher than that of the vehicle ahead. However, we do not impose such a constraint in this work. In other words, the vehicle behind moves slower than the vehicle ahead results in a negative value of TTC, thus the relative distance between these two vehicles will increase. We define the independent variables as listed in Table \ref{Definitions of Independent Variables}. All variables in Table \ref{Definitions of Independent Variables} representing the relative relationship are calculated relative to the ego vehicle. It should be careful to compute $TTC^{\mathrm{lead}}$ in the rude scenario, which is computed by

% TTC_lead
\begin{equation}\label{eq:ttc_lead}
TTC^{\mathrm{lead}} = 
\begin{cases}

\frac{|x^{\mathrm{ego}} - x^{\mathrm{lead}}| - \frac{1}{2} (l^{\mathrm{ego}} + l^{\mathrm{lead}})}{v_{x}^{\mathrm{lead}} - v_{x}^{\mathrm{ego}}}, & \mathrm{if} \ \mathrm{Condition1} \\

\frac{|x^{\mathrm{lead}} - x^{\mathrm{ego}}| - \frac{1}{2} (l^{\mathrm{ego}} + l^{\mathrm{lead}})}{v_{x}^{\mathrm{ego}} - v_{x}^{\mathrm{lead}}}, & \mathrm{if} \ \mathrm{Condition2}

\end{cases}
\end{equation}
where Condition1 is `the lead vehicle is in the left-behind area of the ego vehicle' and Condition2 is `the lead vehicle is in the left-ahead area of the ego vehicle', and $TTC^{\mathrm{lag}}$ is computed by

% TTC_lag
\begin{equation}\label{eq:ttc_lag}
TTC^{\mathrm{lag}} = \frac{|x^{\mathrm{ego}} - x^{\mathrm{lag}}| - \frac{1}{2} (l^{\mathrm{ego}} + l^{\mathrm{lag}})}{v_{x}^{\mathrm{lag}} - v_{x}^{\mathrm{ego}}}
\end{equation}

% Table of Definitions of Independent Variables
\renewcommand\arraystretch{1.5} % default: 1.0
\begin{table}[tb]
	\centering
	\caption{Definitions of Independent Variables}\label{Definitions of Independent Variables}
	\begin{tabular}{p{0.2\columnwidth} p{0.1\columnwidth} p{0.55\columnwidth}}

		% \multicolumn{2}{c}{Independent variables} \\
		\hline \hline
		
		$\Delta {x}^{\mathrm{lead}}$ & [m] & The longitudinal relative distance of the assumptive lead vehicle and ego vehicle \\
		\hline
		
		$\Delta v_{x}^{\mathrm{lead}}$ & [m/s] & The longitudinal relative speed of the assumptive lead vehicle and ego vehicle \\
		\hline
		
		$TTC^{\mathrm{lead}}$ & [s] & The time to collision between the assumptive lead vehicle and ego vehicle \\
		\hline
		
		$v_x^{\mathrm{ego}}$ & [m/s] & The longitudinal speed of the ego vehicle \\
		\hline
		
		$v_y^{\mathrm{ego}}$ & [m/s] & The lateral speed of the ego vehicle \\
		\hline
		
		$\Delta {x}^{\mathrm{lag}}$ & [m] & The longitudinal relative distance of the assumptive lag vehicle and ego vehicle \\
		\hline
		
		$\Delta v_{x}^{\mathrm{lag}}$ & [m/s] & The longitudinal relative speed of the assumptive lag vehicle and ego vehicle \\
		\hline
		
		$TTC^{\mathrm{lag}}$ & [s] & The time to collision between the assumptive lag vehicle and ego vehicle \\
		\hline	\hline
	\end{tabular}
\end{table}

In addition to the \textit{independent} variables defined above, we also need to introduce \textit{dependent} variables that can reflect the procedure of the highway on-ramp merge task. This task requires the ego vehicle to merge onto the main road as soon as possible while ensuring safety. The longitudinal distance of the ego vehicle to the end of the ramp, $\Delta {x}^{\mathrm{end}}$, is used to describe the urgent level. That is, a short distance left increases the urgent level of merge intent. We also define the lateral distance to the lane change boundary line $\Delta {y}^{\mathrm{bdry}}$ to describe how much the task has been completed. So these two indicators are selected as the dependent variables.

\section{Dataset and Data Processing}

\subsection{Real-World Dataset}

It is necessary to adopt real-world driving scenarios as a research basis in order to study human-like maneuvers. The accessible realistic driving datasets include the Next Generation SIMulation (NGSIM) dataset \cite{alexiadis2004next}, the HDD dataset \cite{ramanishka2018toward}, the Argoverse dataset \cite{chang2019argoverse}, the highD dataset \cite{krajewski2018highd}, and the INTERnational, Adversarial and Cooperative moTION (INTERACTION) dataset \cite{zhan2019interaction}. We utilize the INTERACTION dataset for the following reasons: 

\begin{itemize}

\item It includes diversified interactive driving scenarios, such as intersections, roundabouts, and merging scenarios. 

\item In addition to regular driving behaviors and safe operations, highly interactive and complex driving behaviors are also densely contained, such as negotiations, adversarial/irrational decisions, and near-collision maneuvers.

\item It contains well-defined physical information, such as agents' position and speed in longitudinal and lateral directions, the corresponding timestamp with a resolution of $100$ ms, the types of tracked agents (cars or trucks), yaw angle, and the length and width of vehicles.

\end{itemize}

\subsection{Data Preprocessing}

\begin{figure}[t]
\centering
\subfloat[Real scene]{\label{level2.sub.1} \includegraphics[width=0.98\linewidth]{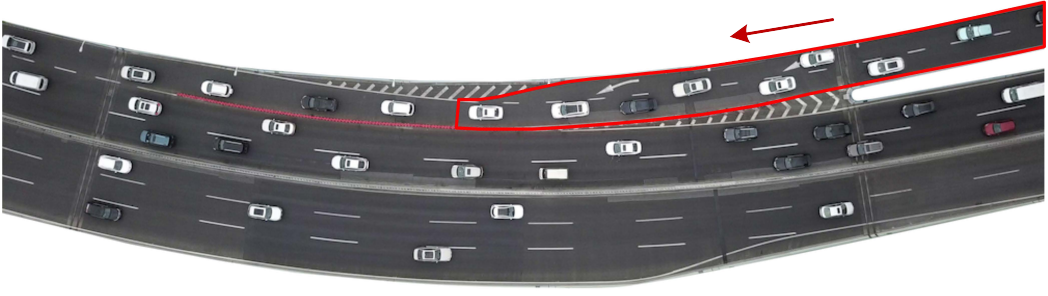}} \\
\subfloat[Data visualization]{\label{level2.sub.2} \includegraphics[width=0.98\linewidth]{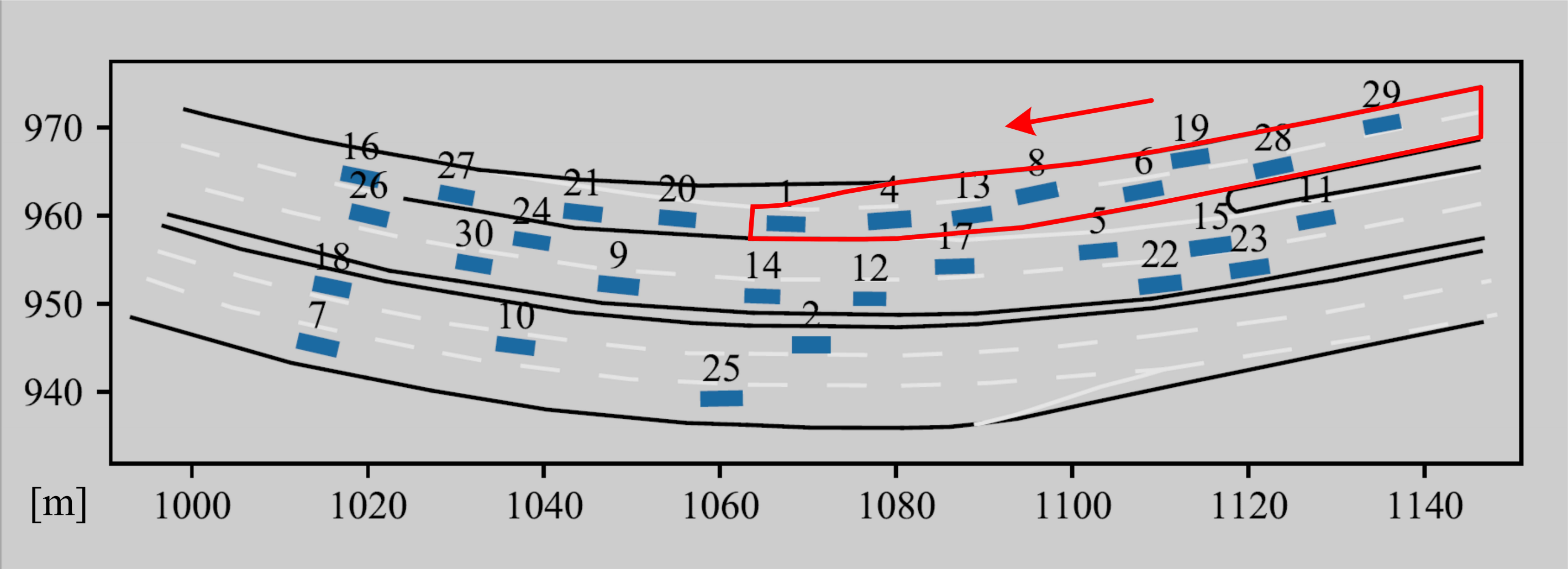}}
\caption{The highway on-ramp merge scenario in the INTERACTION dataset \cite{zhan2019interaction} and the selected local region bounded by the red line.}
\label{fig:INTERACTION dataset}
\end{figure}

The INTERACTION dataset consists of two types of highway merge scenarios across countries: China and Germany. The video length of the Chinese (German) merge scenario is $94.62$ ($37.92$) minutes, which contains $10359$ ($574$) vehicles. The upper two lanes of the Chinese scenario (see Fig. \ref{fig:INTERACTION dataset}) are specially selected because they cover a longer duration and a wider variety of social preferences.

% The highway on-ramp merge scenario in the INTERACTION dataset

The INTERACTION dataset provides a bird-view image of the highway on-ramp scene, as shown in Fig. \ref{fig:INTERACTION dataset}(a). The sub-dataset for each highway on-ramp includes a map file, providing detailed map coordinate information and driving records, including the tracking information of all vehicles, such as vehicle ID, timestamp, vehicle position, and speed. We extracted the coordinates of the boundary between the merge lane and the rightmost highway lane within the selected area based on the map file (\texttt{DR\underline{~}CHN\underline{~}Merging\underline{~}ZS.osm}). Then, we developed a function in Python programming to extract different merging behaviors from the associated tracking data file (\texttt{vehicle\underline{~}tracks\underline{~}$\ast$.csv}) and save them for analysis. This procedure is achieved via FOUR steps:
	
\begin{itemize}
	
\item \textbf{Step 1:} Selecting the merging vehicle. We defined the vehicle as a merging vehicle if the vehicle crossed the boundary from the merge lane to the main road in the selected local region, as shown in Fig.  \ref{fig:INTERACTION dataset}.
	
\item \textbf{Step 2:} Determining the critical moments ($t_s$, $t_m$ and $t_e$) of each merging event. After selecting the merging vehicle via \textbf{Step 1}, we extracted the critical moments (see Section \ref{subsec:cri_moments}) by searching back and forward over the timestamps. 
	
\item \textbf{Step 3:} Labeling the surrounding vehicles paired in the same merging event of the merging vehicle. \textbf{Step 2} allows us to extract all the data in $[t_{s}, t_{e}]$. Here, we mainly consider three involved agents: the ego vehicle, the assumptive lead, and the lag vehicles in the target lane. The longitudinal relative position relationship between the merging vehicle and the surrounding vehicles on the target lane will not change since the end moment $t_e$. Therefore, we selected the closest vehicles to the merging vehicle in the target lane at $t_e$ moment as the lead and lag vehicles. We stored the merge events to ensure that all vehicles included in the event have data records between $t_s$ and $t_e$.
	
\item \textbf{Step 4:} Classifying the extracted merging events according to social preference. We then classified those merging events from \textbf{Step 3} based on the surrounding vehicles (defined in Fig. \ref{fig:Diagram of the typical highway on-ramp merge scenario} and Section \ref{subsec: sur_vehicles}) and social preferences (defined in Section \ref{subsec:social}). The relative position relationship between vehicles allows classifying the merging event as rude or courteous depending on if the lead vehicle is located at the left rear area of the driving direction of the ego vehicle at the moment $t_s$.
\end{itemize}

Finally, 288 rude and 789 courteous merging events were extracted from the selected local scenarios and saved for further analysis. We should note that the three vehicles' longitudinal positional relationship is unchanged in a courteous scenario. In the rude scenario, however, the longitudinal positional relationship will reverse once: At the start moment $t_s$, the lead vehicle is upstream of the traffic flow and acts rudely to force the merging vehicle to yield. Considering this, we calculate the TTC between the merging vehicle and the surrounding vehicle by using the positional relationship in Equations (\ref{eq:ttc_lead}) and (\ref{eq:ttc_lag}). Based on the selected variables in Section \ref{subsec:variable}, we calculated the independent variables and dependent variables. We then divided the extracted data into three groups for each scenario according to the three predefined critical moments to analyze interactions over the whole procedure with different social preferences.

\subsection{Methods}

The Analysis of Variance (ANOVA) is a mature method for analyzing the significant level of variables through the significance test \cite{bray1985multivariate}. It can deal with mixture analysis in which the independent variable is qualitative while the dependent variable is quantitative. To meet the requirements of ANOVA, for both merging scenarios, we divided all the independent variables at different moments into two groups according to their median. In this way, the grouped independent variables and quantitative dependent variables were analyzed by ANOVA (using the SPSS software) to obtain the significance level of independent variables at different times for different merge behaviors.

\section{Result Analysis and Discussion}

\subsection{Analysis of Independent Variables}

\begin{figure}[tb]
\centering
\subfloat[]{\label{level3.sub.1} \includegraphics[width=0.96\linewidth]{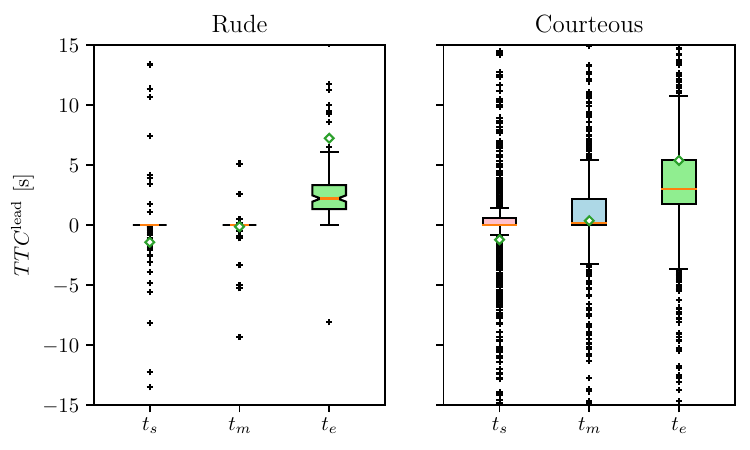}}\\
\subfloat[]{\label{level3.sub.2} \includegraphics[width=0.96\linewidth]{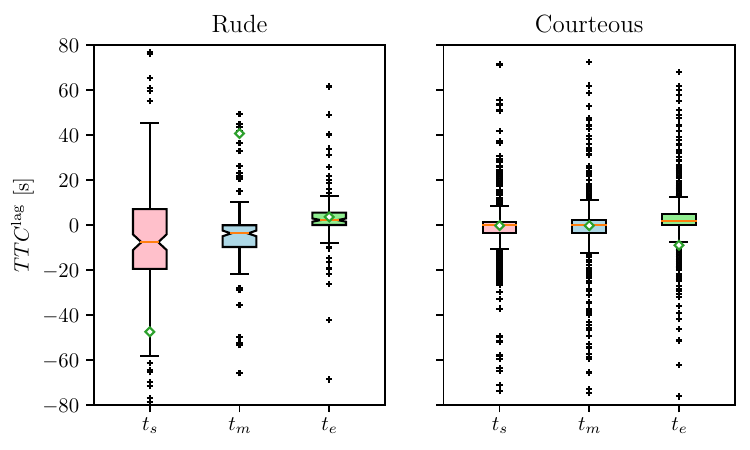}}
\caption{Statistical results of (a)
$TTC^{\mathrm{lead}}$ and (b) $TTC^{\mathrm{lag}}$ at the three critical moments ($t_{s}$, $t_{m}$, and $t_{e}$) in rude and courteous scenarios.}
\label{fig:ttc}
\end{figure}

Figs. \ref{fig:independent_variables} and \ref{fig:ttc} show the statistical results of the independent variables with rude and courteous social preferences at the start ($t_s$), middle ($t_m$), and end ($t_e$) moments. A comprehensive comparison of these variables reflects the changes in decision-making during the merging process. Note that the vehicles on the selected highway and associated ramp move left (see Fig \ref{fig:INTERACTION dataset} (a)), so the absolute values of the longitudinal coordinates gradually decrease (see Fig. \ref{fig:INTERACTION dataset} (b)), indicating that the speed of all the selected vehicles is negative. In what follows, we will discuss and analyze the dynamic merging process of the ego vehicle when interacting with the different social preferences (i.e., rude and courteous) of the surrounding vehicles.

\subsubsection{Interactions with rude social preferences}
\begin{figure}[t]
\centering
\includegraphics[width=\linewidth]{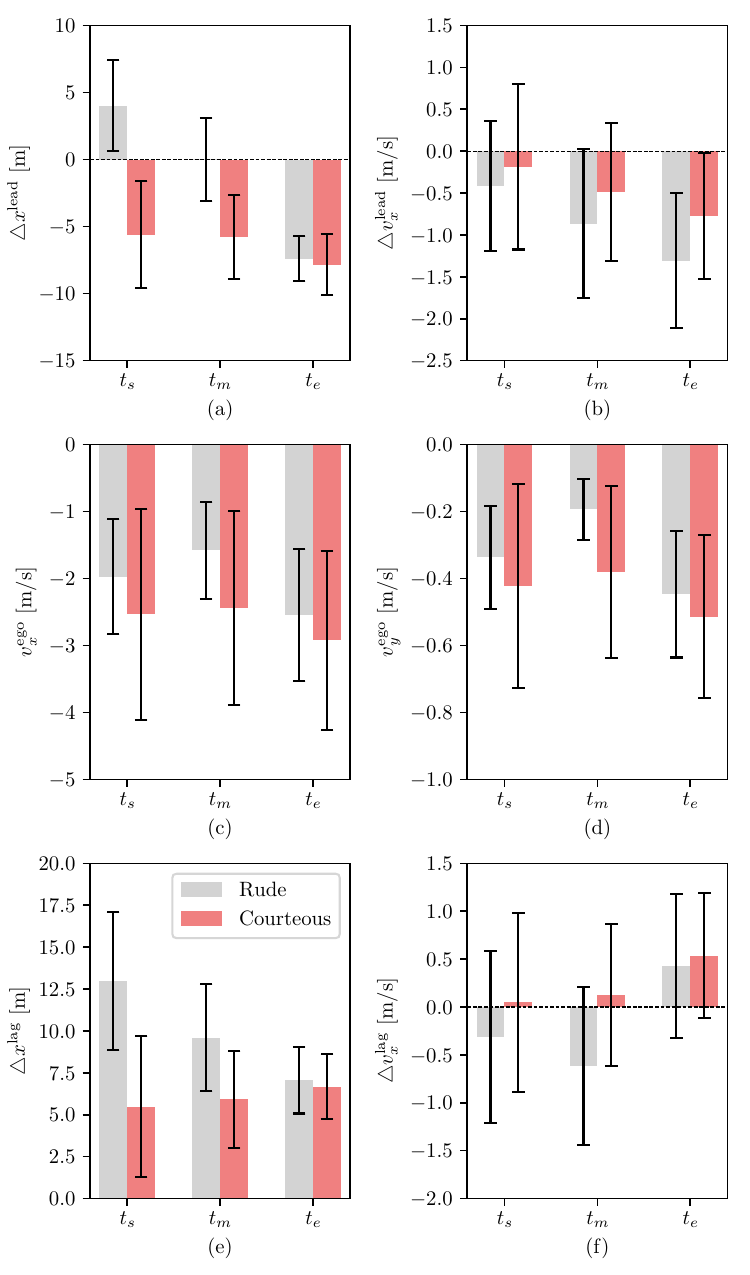}
\caption{Statistical results of independent variables at the three critical moments ($t_{s}$, $t_{m}$, and $t_{e}$) in the rude and courteous scenarios.}
\label{fig:independent_variables}
\end{figure}

For the relative position between the ego vehicle and its target vehicles, the grey bars' values in Fig. \ref{fig:independent_variables}(a) show that the relative distance ($\Delta x^{\mathrm{lead}}$) changes from positive to negative over the merging process. It indicates that the lead vehicle acts rudely and passes the merging vehicle in the longitudinal direction, as illustrated in Fig. \ref{fig:Diagram of the typical highway on-ramp merge scenario}(b). This is consistent with the definition of the rude scenario: the human driver in the lead vehicle has a competitive social preference and is more self-centered, and thus does not allow the ego vehicle to merge into the gap ahead of it. The mean value of $\Delta {x}^{\mathrm{lead}}$ at $t_m$ is approximately equal to zero, which means the lead vehicle almost drives side-by-side with the ego vehicle. 

% rude
% vx_ahead:  -2.38, -2.44, -3.86
%    vx:     -1.97, -1.58, -2.55
% vx_behind: -2.28, -2.20, -2.12

The gray bars in Fig. \ref{fig:independent_variables}(c) and (d) show that $v_x^{\mathrm{ego}}$ and $v_y^{\mathrm{ego}}$ have identical speed trends: decreasing first and increasing. At the middle moment, both the longitudinal and lateral speeds of the ego vehicle decrease to the lowest absolute value because the ego vehicle needs to understand the behavior of surrounding vehicles and avoid collisions. Moreover, $\Delta v_{x}^{\mathrm{lead}}$ in Fig. \ref{fig:independent_variables}(b) changes significantly from $t_s$ to $t_m$ because the ego vehicle changes its longitudinal speed a lot while the lead vehicle adjusts its speed slightly. Also, at the middle moment $t_m$, $\Delta v_{x}^{\mathrm{lead}}$ is generally less than zero, which indicates that the lead vehicle is moving faster than the ego vehicle. After $t_m$, both the ego and lead vehicles will gradually increase their speed by about 60\% to be consistent with the traffic flow. However, the average value of $\Delta v_{x}^{\mathrm{lag}}$ (the gray bars in Fig. \ref{fig:independent_variables}(f)) keeps decreasing slightly from $t_s$ to $t_e$. The above analysis indicates that the lead vehicle behaves rudely and will give a high priority to keep moving forward at a near-constant speed in the merging process. The lag vehicle keeps decelerating throughout the process to leave a gap for the merging vehicle to cut in.

% independent_variables

% Figure of TTC 

\subsubsection{Interactions with courteous social preferences}
Regarding the position of the involved vehicles, the red bars in Fig. \ref{fig:independent_variables}(a) and (e) show that the values of $\Delta {x}^{\mathrm{lead}}$ and $\Delta {x}^{\mathrm{lag}}$ are negative and positive for all three moments. It indicates that the ego vehicle always stays in the middle between the assumptive lead and lag vehicles during the merging process, as shown in Fig. \ref{fig:Diagram of the typical highway on-ramp merge scenario}(a). However, the averages of $\Delta {x}^{\mathrm{lead}}$ and $\Delta {x}^{\mathrm{lag}}$ increase along with the merging process. It reveals that the lag vehicle acts courteously by actively adjusting its speed to ensure an adequate safety gap ahead, thus allowing the ego vehicle to cut in. 

% courteous
% vx_ahead:  -2.73, -2.93, -3.69
%    vx:     -2.54, -2.44, -2.92
% vx_behind: -2.49, -2.32, -2.38

In terms of speed, both $v_x^{\mathrm{ego}}$ and $v_y^{\mathrm{ego}}$ decrease first and increase then, which have the same interactions in a rude scenario. However, due to the courteous yield behavior of the lag vehicle, the duration of deceleration caused by the merge duration is shorter than in the rude scenario, causing more minor speed changes from $t_s$ to $t_e$. At the end moment $t_e$, all the averages of $\Delta v_{x}^{\mathrm{lead}}$ are negative, while almost all the averages of $\Delta v_{x}^{\mathrm{lag}}$ are positive, which means that the ego vehicle will actively decelerate to keep a safe distance ahead, while the courteous human driver of the lag vehicle will also slow down after detecting the merging intention and then yield. The ego vehicle can interactively respond to the courteous behavior of the lag vehicle. Hence, although both the ego and lag vehicles suffer a speed loss, the speed changes over the three critical moments are small because the game period is shorter than in a rude scenario.

\subsubsection{Comparisons}
Fig. \ref{fig:independent_variables}(a) and (e) reveal that although the merging processes with rude and courteous social preferences are quite different, their absolute mean value and variance of $\Delta {x}^{\mathrm{lead}}$ for $t_e$ are almost consistent. The same conclusion can be obtained for $\Delta {x}^{\mathrm{lag}}$. Besides, the mean values of $\Delta {x}^{\mathrm{lead}}$ and $\Delta {x}^{\mathrm{lag}}$ in the courteous scenario are almost equal to those in the rude scenario at $t_{e}$. Moreover, both $\Delta {x}^{\mathrm{lead}}$ and $\Delta {x}^{\mathrm{lag}}$ obtain the lowest value at $t_e$, which indicates that human drivers will adjust their relative position and speed to achieve the merging task by finally keeping a relatively safe gap (about $5\sim 7$ m). However, the changes in $\Delta {x}^{\mathrm{lead}}$ and $\Delta {x}^{\mathrm{lag}}$ after $t_e$ are the opposite: $\Delta v_{x}^{\mathrm{lead}}$ at the end moment $t_e$ is negative while most of the $\Delta v_{x}^{\mathrm{lag}}$ is positive. That is, the gaps (i.e., $|\Delta {x}^{\mathrm{lead}}|$ and $|\Delta {x}^{\mathrm{lag}}|$) will continue to increase after merging. In addition, when interacting with a rude surrounding driver, $\Delta v_{x}^{\mathrm{lead}}$ obtains the highest average value, indicating that the surrounding driver with a rude preference tends to accelerate after passing over the ego vehicle.

By comparing $v_x^{\mathrm{ego}}$ with $v_y^{\mathrm{ego}}$ in Fig. \ref{fig:independent_variables}(c) and (d), we can see that the ego vehicle will actively slow down in both longitudinal and lateral directions to ensure a safe merge, while the speed change of $v_y^{\mathrm{ego}}$ is particularly apparent, especially when interacting with a rude surrounding driver. Besides, Fig. \ref{fig:independent_variables}(b) and (f) reveal that the changes of $\Delta v_{x}^{\mathrm{lead}}$ and $\Delta v_{x}^{\mathrm{lag}}$ from $t_{s}$ to $t_e$ in the courteous scenario are more stable than in the rude scenario. 

Fig. \ref{fig:ttc} displays the statistical results of $TTC^{\mathrm{lead}}$ and $TTC^{\mathrm{lag}}$, and their changes are consistent with those indicated by position and speed. One interesting finding is that $TTC^{\mathrm{lag}}$ changes more slightly than $TTC^{\mathrm{lead}}$, while either variable in the courteous scenario changes more steadily than that in the rude scenario.

% Remaining longitudinal distance to the end of merge lane
\begin{figure}[t]
\centering
\includegraphics[width=\linewidth]{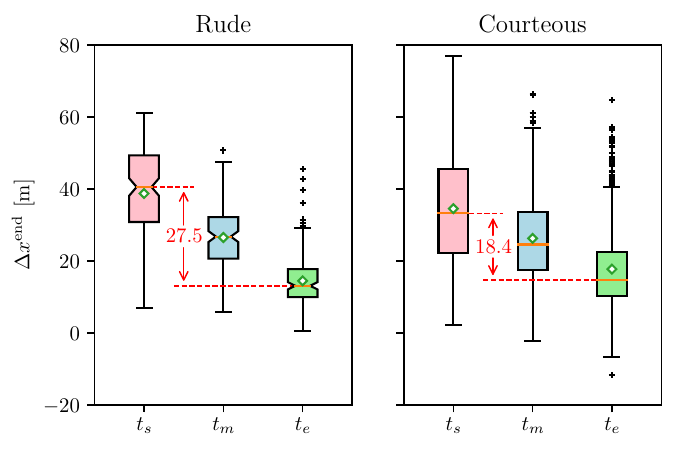}
\caption{Statistical results of $\Delta {x}^{\mathrm{end}}$, the remaining longitudinal distance to the end of the merge lane at the three critical moments ($t_{s}$, $t_{m}$, and $t_{e}$) in rude and courteous scenarios.}
\label{fig:remaining_distance}
\end{figure}

% Table of Chinese Significance Analysis
\renewcommand\arraystretch{1.5} % default: 1.0
\begin{table*}[t]
	\centering
	\caption{The Chinese Results of Significance Analysis ($p$-value) on the Selected Variables at the Three Critical Moments ($t_{s}$, $t_{m}$, and $t_{e}$) in Rude and Courteous Scenarios (with $^{\ast} p<0.05$; $^{\ast\ast}p<0.01$; $^{\ast\ast\ast}p<0.001$)}\label{Chinese Significance Analysis}
	% \begin{tabular}{p{0.08\textwidth} p{0.08\textwidth} p{0.08\textwidth} p{0.08\textwidth} p{0.08\textwidth} p{0.08\textwidth} p{0.08\textwidth} p{0.08\textwidth} p{0.08\textwidth}}
	\begin{tabular}{ c | c | c | c c c c c c c c }

		\hline \hline
		
		&  &  & $\Delta {x}^{\mathrm{lead}}$ & $\Delta v_{x}^{\mathrm{lead}}$ & $TTC^{\mathrm{lead}}$ & $v_x^{\mathrm{ego}}$ & $v_y^{\mathrm{ego}}$ & $\Delta {x}^{\mathrm{lag}}$ & $\Delta v_x^{\mathrm{lag}}$ & $TTC^{\mathrm{lag}}$ \\
		
		\hline \hline
		
		% x_end
		\multirow{6}*{$\Delta {x}^{\mathrm{end}}$} & \multirow{3}*{\rotatebox{90}{Rude}} & $t_s$ & \cellcolor{lightgray!50}-- & -- & -- & $0.003^{\ast\ast}$ & $<0.001^{\ast\ast\ast}$ & \cellcolor{lightgray!50}-- & -- & -- \\
		\cline{3-11}
		~ & ~ & $t_m$ & \cellcolor{lightgray!50}-- & \cellcolor{orange!90}$<0.001^{\ast\ast\ast}$ & $0.042^{\ast}$ & $0.001^{\ast\ast}$ & $<0.001^{\ast\ast\ast}$ & \cellcolor{lightgray!50}-- & $0.032^{\ast}$ & $0.024^{\ast}$ \\
		\cline{3-11}
		~ & ~ & $t_e$ & \cellcolor{lightgray!50}-- & -- & $0.028^{\ast}$ & -- & -- & \cellcolor{lightgray!50}-- & \cellcolor{orange!90}$0.008^{\ast\ast}$ & -- \\
		
		\cline{2-11} \cline{2-11}
		
		& \multirow{3}*{\rotatebox{90}{Courteous}} & $t_s$ & $0.003^{\ast\ast}$ & \cellcolor{orange!30!red}$<0.001^{\ast\ast\ast}$ & -- & $<0.001^{\ast\ast\ast}$ & $<0.001^{\ast\ast\ast}$ & -- & -- & $0.001^{\ast\ast}$ \\
		\cline{3-11}
		~ & ~ & $t_m$ & $0.001^{\ast\ast}$ & \cellcolor{orange!30!red}$<0.001^{\ast\ast\ast}$ & $0.005^{\ast\ast}$ & $<0.001^{\ast\ast\ast}$ & $<0.001^{\ast\ast\ast}$ & \cellcolor{orange!30!red}$<0.001^{\ast\ast\ast}$ & -- & -- \\
		\cline{3-11}
		~ & ~ & $t_e$ & -- & \cellcolor{orange!30!red}$<0.001^{\ast\ast\ast}$ & -- & $<0.001^{\ast\ast\ast}$ & $<0.001^{\ast\ast\ast}$ & -- & $0.012^{\ast}$ & $0.014^{\ast}$ \\
		
		\hline \hline
		
		% y_boundary
		\multirow{6}*{$\Delta {y}^{\mathrm{bdry}}$} & \multirow{3}*{\rotatebox{90}{Rude}} & $t_s$ & \cellcolor{lightgray!50}-- & -- & -- & -- & -- & \cellcolor{lightgray!50}-- & -- & -- \\
		\cline{3-11}
		~ & ~ & $t_m$ & \cellcolor{lightgray!50}-- & -- & $0.001^{\ast\ast}$ & $0.002^{\ast\ast}$ & $<0.001^{\ast\ast\ast}$ & \cellcolor{lightgray!50}-- & -- & -- \\
		\cline{3-11}
		~ & ~ & $t_e$ & \cellcolor{lightgray!50}-- & -- & -- & -- & -- & \cellcolor{lightgray!50}-- & -- & -- \\
		
		\cline{2-11} \cline{2-11}
		
		& \multirow{3}*{\rotatebox{90}{Courteous}} & $t_s$ & -- & -- & -- & $<0.001^{\ast\ast\ast}$ & $0.001^{\ast\ast}$ & -- & -- & -- \\
		\cline{3-11}
		~ & ~ & $t_m$ & -- & -- & $0.019^{\ast}$ & $<0.001^{\ast\ast\ast}$ & $<0.001^{\ast\ast\ast}$ & -- & \cellcolor{orange!30!red}$<0.001^{\ast\ast\ast}$ & -- \\
		\cline{3-11}
		~ & ~ & $t_e$ & -- & -- & -- & -- & -- & $0.038^{\ast}$ & -- & -- \\

		\hline \hline
	\end{tabular}
\end{table*}

\subsection{Analysis of Dependent Variables}

The above analysis is only based on the independent variables. Here, we will analyze the dependent variables $\Delta {x}^{\mathrm{end}}$ and $\Delta {y}^{\mathrm{bdry}}$. The definitions of the three critical moments (in Section \ref{subsec:cri_moments}) make the distribution of $\Delta {y}^{\mathrm{bdry}}$ the same at each moment. Therefore, we will mainly show and discuss $\Delta {x}^{\mathrm{end}}$ (Fig. \ref{fig:remaining_distance}), instead $\Delta {y}^{\mathrm{bdry}}$ here. 

\begin{figure}[tb]
\centering
\subfloat[Rude]{\label{level4.sub.1} \includegraphics[width=0.98\linewidth]{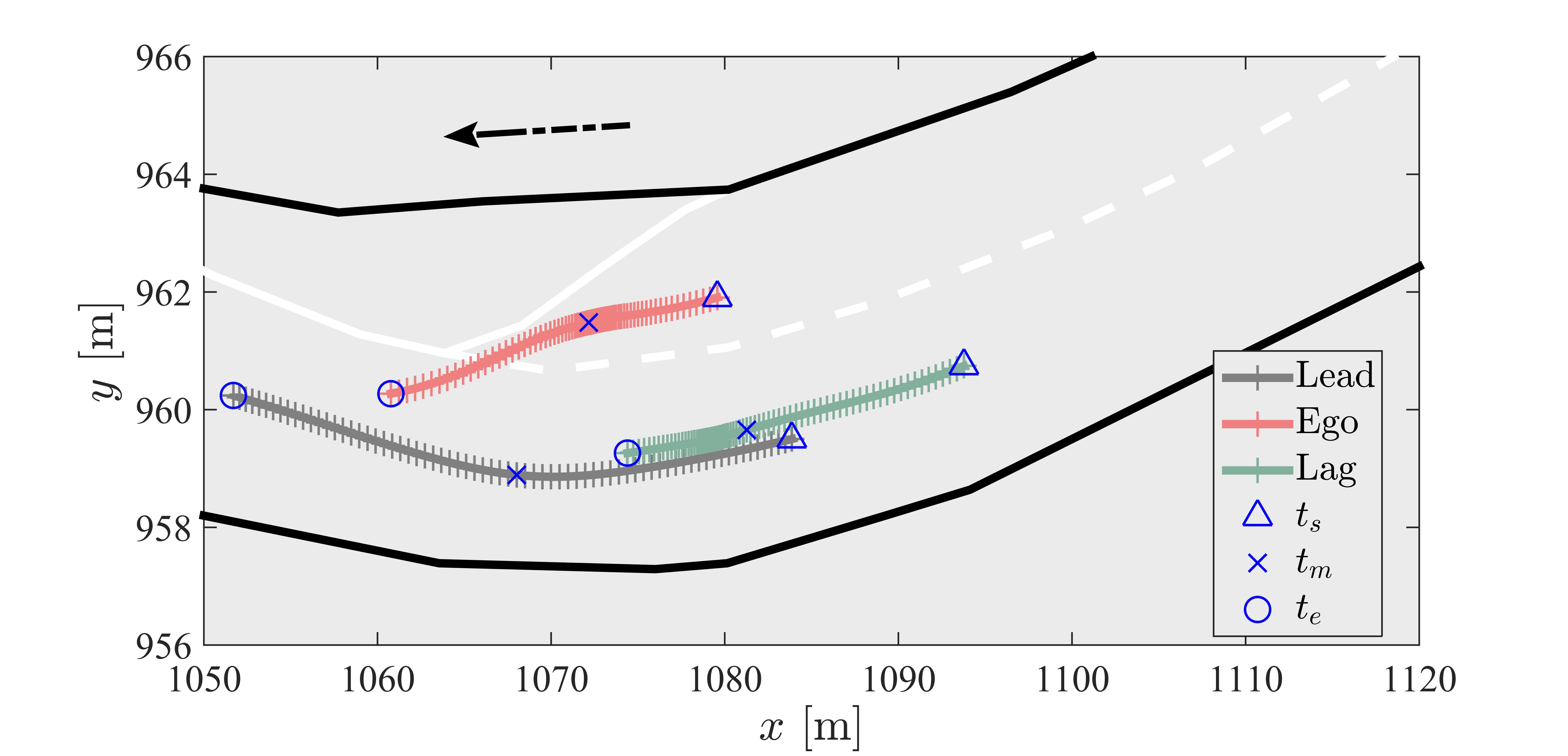}}\\
\subfloat[Courteous]{\label{level4.sub.2} \includegraphics[width=0.98\linewidth]{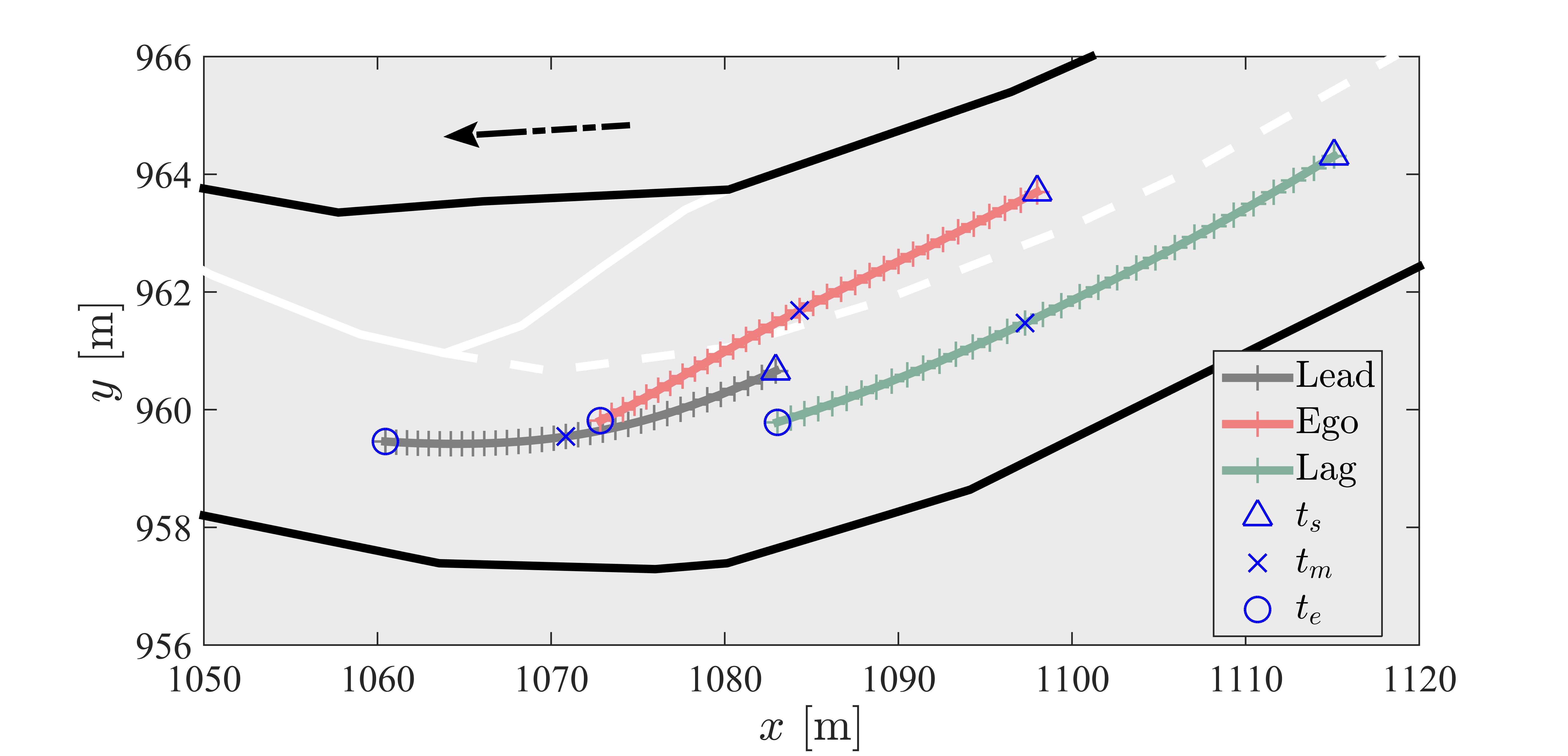}}
\caption{Illustration of trajectory examples in (a) rude and (b) courteous scenarios. Dense (sparse) `+' markers indicate a low (high) speed.}
\label{fig:trajectory}
\end{figure}

Fig. \ref{fig:remaining_distance} reveals that the median and mean values of $\Delta {x}^{\mathrm{end}}$ at moments $t_m$ and $t_e$ in the two social scenarios are not significantly different. The longitudinal driving distance of the ego vehicle throughout the merging process when interacting with a rude surrounding driver ($27.5$ m) is higher than with a courteous surrounding driver ($18.4$ m). It underlies that \textit{courteous social interaction may improve traffic efficiency}. Moreover, the rude merging events occur more often near the end of the ramp (see Fig. \ref{fig:remaining_distance} and Fig. \ref{fig:trajectory}). When the ego vehicle approaches the end of the ramp gradually, its merging intention becomes strong. The ego vehicle continues to forcibly merge even if the surrounding environmental conditions do not guarantee safety. The defensive actions of the merging vehicles may also trigger the adversarial or competitive responses of the surrounding drivers.

Fig. \ref{fig:trajectory} illustrates the difference between the ego vehicle's trajectory in different interaction scenarios. In the \textit{rude} merging interaction, the surrounding vehicles competitively force the ego vehicle to drive parallelly to the boundary line along the ramp at a low speed (maybe close to zero at $t_m$). Once overtaken by the lead vehicle, the ego vehicle will rapidly increase longitudinal and lateral speed to complete the merging task quickly and safely. Therefore, the ego vehicle first moves ahead straightly at a slight angle to the boundary and then merges into the traffic smoothly by following a curved trajectory. In the \textit{courteous} merging interaction, the surrounding vehicles will give way to the merging vehicle. Thus, the ego vehicle slightly changes the longitudinal speed, increases lateral speed first, and decreases smoothly. Therefore, the ego vehicle's trajectory consists of two smooth, continuous curves with a slight curvature.

On the other hand, Fig. \ref{fig:remaining_distance} shows that $\Delta {x}^{\mathrm{end}}$ is almost always positive in the rude scenario. It indicates that the merging behaviors are completed before the end of the ramp (corresponding to the white diversion solid line on the left side of the red box in Fig. \ref{fig:INTERACTION dataset}(a)). However, $\Delta {x}^{\mathrm{end}}$ obtains some negative values in the courteous scenario. It indicates that parts of the merging behavior are completed after the end of the ramp by driving through the white diversion solid line area, which is usually forbidden to pass. There are two main reasons for this phenomenon: 1) The surrounding traffic conditions in the target lane do not meet the merging conditions, and 2) The ahead gap is large enough for merging. Besides, the higher the ego vehicle's speed is, the more likely this phenomenon will occur. The ego vehicle courteously merges into the target lane at the cost of violating traffic rules, and thus, it gains a shorter merging time and has a low impact on the subsequent ramp traffic flow.

\subsection{Significance Analysis of Variables}

Until this section, the merging process with congested traffic was analyzed based on independent and dependent variables. This section focuses on the significance analysis of variables over time and space and their differences under different social interaction scenarios, which corresponds to the proposed three questions in Section \ref{sec:introduction}. The $p$-value is a random variable derived from the distribution of the test statistic used to analyze a data set. In this work, the significance level was set at $0.05$, for which a $p$-value less than $0.05$ is considered statistically significant: the smaller the $p$-value, the more significant the variable. The results of significance analysis on the selected variables are listed in Table \ref{Chinese Significance Analysis}. The significance analysis of $\Delta {y^{\mathrm{bdry}}}$ is consistent with that of $\Delta {x}^{\mathrm{end}}$ although with small differences. Analyzing the results for the two dependent variables comprehensively, we can get three key conclusions as follows.

\textbf{1) The social preferences of the surrounding vehicles impact the variable selection of the ego vehicle to make a decision.} 
When interacting with a rude surrounding driver, the ego driver would mainly rely on the relative velocity (highlighted as orange) rather than the relative distance at any moment (highlighted as gray). In other words, the ego vehicle will not use the full state of the environment to make decisions when interacting with a rude surrounding driver in merge scenarios. Conversely, when interacting with a courteous surrounding driver, the ego vehicle would select \textit{all} the selected independent variables to make decisions during the whole merge process. Specifically, the relative distance and relative speed are the most significant (highlighted as red) for the ego vehicle but with different significance levels over the three critical moments. In other words, the ego vehicle will use more environmental states to make decisions in a courteous scenario than in a rude scenario. However, $v_x^{\mathrm{ego}}$ and $v_y^{\mathrm{ego}}$ are significant for the ego vehicle (except $t_e$ in the rude scenario) in both courteous and rude scenarios.

\textbf{2) Variable saliency varies over the merging process.} The changes in variable saliency in the merging process in different scenarios are different.

In the rude scenario, at the moment
\begin{itemize}

\item $t_s$: $v_x^{\mathrm{ego}}$ ($p=0.003$) and $v_y^{\mathrm{ego}}$ ($p<0.001$) are the only significant variables. The ego vehicle attempts to merge by triggering an interaction with its surrounding vehicles at the initial stage of merging by only evaluating their longitudinal and lateral speed. 

\item $t_m$: The $\Delta v_{x}^{\mathrm{lead}}$ is also the most significant variable (with $p<0.001$) besides $v_x^{\mathrm{ego}}$ and $v_y^{\mathrm{ego}}$ (both with $p<0.001$). Moreover, the merging decision of the ego vehicle also slightly relies on $TTC^{\mathrm{lead}}$ ($p=0.042$), $\Delta v_x^{\mathrm{lag}}$ ($p=0.032$) and $TTC^{\mathrm{lag}}$ ($p=0.024$). The slight dependence on $\Delta v_x^{\mathrm{lag}}$ and $TTC^{\mathrm{lag}}$ indicates that the ego vehicle needs to take care of the intention of the lag vehicle while interacting with the lead vehicle to judge the social preference of the lag vehicle. 

\item $t_e$: The ego vehicle makes decisions by mainly relying on $\Delta v_x^{\mathrm{lag}}$ ($p=0.008$) while slightly relying on $TTC^{\mathrm{lead}}$ ($p=0.028$). The intuitive explanation of this is that the ego vehicle still needs to pay more attention to the lag vehicle's behavior to ensure safety while following the lead vehicle because the ego vehicle almost completes the merging task at this moment. 

\end{itemize}

In the courteous scenario, at the moment
\begin{itemize}
    
\item $t_s$: The ego vehicle makes decisions mainly based on its relative speed to the lead vehicle, $\Delta v_{x}^{\mathrm{lead}}$ ($p<0.001$), while slightly depending on the distance to the lead vehicle $\Delta {x}^{\mathrm{lead}}$ ($p=0.003$) and the risk level with the lag vehicle $TTC^{\mathrm{lag}}$ ($p=0.001$).

\item $t_m$: The ego vehicle takes actions depending mainly on $\Delta {x}^{\mathrm{lag}}$ ($p<0.001$) and $\Delta v_x^{\mathrm{lag}}$ ($p<0.001$), while slightly on $\Delta {x}^{\mathrm{lead}}$ ($p=0.001$) and $TTC^{\mathrm{lead}}$ ($p=0.005$).

\item $t_e$: The ego vehicle will make decisions by using, but not significantly, $\Delta {x}^{\mathrm{lag}}$ ($p=0.038$), $\Delta v_x^{\mathrm{lag}}$ ($p=0.012$) and $TTC^{\mathrm{lag}}$ ($p=0.014$). One more interesting finding is that the ego vehicle would put its secondary-attention from the lead vehicle to the lag vehicle. 

\end{itemize}

\textbf{3) Drivers always rely on certain basis variables during the whole merging procedure.} For example, when interacting with a courteous driver, $\Delta v_x^{\mathrm{lead}}$ ($p<0.001$), $v_x^{\mathrm{ego}}$ ($p<0.001$) and $v_y^{\mathrm{ego}}$ ($p<0.001$) are always the most significant variables. The dominant $\Delta v_x^{\mathrm{lead}}$ indicates that the ego vehicle will keep paying attention to its relative speed to the lead vehicle, to make a timely response to contextual changes. 

Based on the above analysis, we summarize the significant levels of each independent variable over the whole merging process while interacting with the rude and courteous surrounding vehicles in Table \ref{table:Chinese summary}. Three significant levels are marked with dark gray ($p<0.001$), gray ($p<0.01$), and light gray ($p<0.05$).

% Table - summary of Chinese significant variables
\renewcommand\arraystretch{1.5} % default: 1.0
\begin{table}[t]
	\centering
	\caption{The Chinese statistics of significant variables at different moments for the rude and courteous scenario (Dark background color represents a high significance level of variables)}\label{table:Chinese summary}

	\begin{tabular}{ c c c | c c c }

		\hline \hline

		\multicolumn{3}{c|}{Rude} & \multicolumn{3}{c}{Courteous} \\
		
		\hline
		
		$t_s$ & $t_m$ & $t_e$ & $t_s$ & $t_m$ & $t_e$ \\
		
		\hline
		
		\cellcolor{lightgray!50}$v_y^{\mathrm{ego}}$ & \cellcolor{lightgray!50}$v_y^{\mathrm{ego}}$ & \cellcolor{lightgray!25}$\Delta v_x^{\mathrm{lag}}$ & \cellcolor{lightgray!50}$v_y^{\mathrm{ego}}$ & \cellcolor{lightgray!50}$v_y^{\mathrm{ego}}$ & \cellcolor{lightgray!50}$v_y^{\mathrm{ego}}$ \\
		
		\hline
		
		\cellcolor{lightgray!25}$v_x^{\mathrm{ego}}$ & \cellcolor{lightgray!50}$\Delta v_{x}^{\mathrm{lead}}$ & \cellcolor{lightgray!5}$TTC^{\mathrm{lead}}$ & \cellcolor{lightgray!50}$v_x^{\mathrm{ego}}$ & \cellcolor{lightgray!50}$v_x^{\mathrm{ego}}$ & \cellcolor{lightgray!50}$v_x^{\mathrm{ego}}$ \\
		
		\hline
		
		& \cellcolor{lightgray!25}$v_x^{\mathrm{ego}}$ & & \cellcolor{lightgray!50}$\Delta v_{x}^{\mathrm{lead}}$ & \cellcolor{lightgray!50}$\Delta v_{x}^{\mathrm{lead}}$ & \cellcolor{lightgray!50}$\Delta v_{x}^{\mathrm{lead}}$ \\
		
		\hline
		
		& \cellcolor{lightgray!5}$TTC^{\mathrm{lead}}$ & & \cellcolor{lightgray!25}$\Delta {x}^{\mathrm{lead}}$ & \cellcolor{lightgray!50}$\Delta {x}^{\mathrm{lag}}$ & \cellcolor{lightgray!5}$\Delta {x}^{\mathrm{lag}}$ \\
		
		\hline
		
		& \cellcolor{lightgray!5}$\Delta v_x^{\mathrm{lag}}$ & & \cellcolor{lightgray!25}$TTC^{\mathrm{lag}}$ & \cellcolor{lightgray!50}$\Delta v_x^{\mathrm{lag}}$ & \cellcolor{lightgray!5}$\Delta v_x^{\mathrm{lag}}$ \\
		
		\hline
		
		& \cellcolor{lightgray!5}$TTC^{\mathrm{lag}}$ & & & \cellcolor{lightgray!25}$\Delta {x}^{\mathrm{lead}}$ & \cellcolor{lightgray!5}$TTC^{\mathrm{lag}}$ \\
		
		\hline
		
		& & & & \cellcolor{lightgray!25}$TTC^{\mathrm{lead}}$ & \\
		
		\hline \hline
	\end{tabular}
\end{table}

% % Table - summary of German significant variables
% \renewcommand\arraystretch{1.5} % default: 1.0
% \begin{table}[t]
% 	\centering
% 	\caption{The German statistics of significant variables at different moments for the rude and courteous scenario (Dark background color represents a high significance level of variables)}\label{table:German summary}

% 	\begin{tabular}{ c c c }

% 		\hline \hline

% 		\multicolumn{3}{c}{Courteous} \\
		
% 		\hline
		
% 		$t_s$ & $t_m$ & $t_e$ \\
		
% 		\hline
		
% 		\cellcolor{lightgray!50}$v_y^{\mathrm{ego}}$ & \cellcolor{lightgray!50}$v_y^{\mathrm{ego}}$ & \cellcolor{lightgray!50}$v_y^{\mathrm{ego}}$ \\
		
% 		\hline
		
% 		\cellcolor{lightgray!50}$v_x^{\mathrm{ego}}$ & \cellcolor{lightgray!50}$v_x^{\mathrm{ego}}$ & \cellcolor{lightgray!50}$v_x^{\mathrm{ego}}$ \\
		
% 		\hline
		
% 		\cellcolor{lightgray!25}$\Delta x^{\mathrm{lead}}$ & \cellcolor{lightgray!25}$\Delta x^{\mathrm{lead}}$ & \cellcolor{lightgray!25}$\Delta x^{\mathrm{lead}}$ \\
		
% 		\hline

% 		& \cellcolor{lightgray!5}$TTC^{\mathrm{lead}}$ & \\
		
% 		\hline \hline
% 	\end{tabular}
% \end{table}

\subsection{Further Discussions}

\subsubsection{Potential Applications}
The multi-dimensional states of the complex environment can overwhelm human insights and analysis. Even in a simple real-world autonomous task, the sensory devices receive a large amount of information, but most of it is useless for task execution \cite{langdon2019uncovering}. Hence, it is crucial to know what information guides humans to make decisions when interacting in a multivariate environment. Moreover, identifying the task-related variables and their significance over time can help advance learning technologies, such as reinforcement learning \cite{song2020learning, niv2019learning}. However, existing research on learning algorithms for the highway on-ramp merge neglects the differences in each variable's contribution to the task execution over space and time \cite{nishi2019merging, bouton2019cooperation}. Therefore, the results of variable significance could be helpful for weighted variable selection to improve model performance \cite{wang2021uncovering}. Besides, the fundamental findings in \cite{karimi2020cooperative} and \cite{guo2020merging} indicate that our findings could help understand and model the interactive objects of autonomous driving with different social preferences and analyze the influence of human drivers' social preferences on the overall traffic flow.

\subsubsection{Influences of Traffic Conditions}
This paper mainly focuses on congested traffic conditions. The vehicle speed is higher in free-flowing traffic conditions, and the gaps between vehicles are larger. The conclusions obtained might not be suitable for the merge task in a free-flow traffic condition due to the differences in merge location and space/time gap \cite{daamen2010empirical}. Therefore, further investigation under different traffic flow conditions is needed in future work.

\subsubsection{Traffic Rules and Driving Habits in Different Countries}

This paper draws conclusions only based on the data collected from China. There are some differences in traffic rules and driving habits across different countries. Drivers' resistance or tolerance to competitive merging interactions in different countries might differ from each other, affecting the selection and significance of significant variables during the merging process. Thus, the social interaction analysis of drivers' merging behavior needs to be conducted based on more diverse datasets in future work.

\subsubsection{Types of Vehicles}
This paper mainly focused on the interactions between cars without considering other vehicles, such as trucks. However, the type of vehicle could influence the lane-change decisions of humans \cite{moridpour2010modeling} and their preferences for rude and courteous. For example, the driver of a passenger car usually leaves ample space ahead when interacting with a fully-loaded truck. Therefore, the influence of the type of vehicle on variable selection and social interactions will be considered in future work.

\section{Conclusion}

This paper provided insights into the influence of the social preferences of surrounding vehicles on merging vehicles' decisions over time and space. We defined three critical moments of the highway on-ramp merge to describe the dynamic merging process. Then, we specified two typical interaction scenarios (i.e., rude and courteous) based on the social preferences of the surrounding vehicles. Further, the selected independent and dependent variables were analyzed based on the INTERACTION dataset with the ANOVA approach. Finally, two fundamental mechanisms for merging tasks at highway on-ramps with congested traffic have been obtained: 

\begin{enumerate}[label=\protect\textcircled{\small\arabic*}]
\item The social preferences of the surrounding vehicles impact the variable selection of the ego vehicle when making decisions. 
\item The variable saliency is not constant; it varies over the merging process with different social preferences in rude and courteous. 
%\item Human drivers always rely on certain fundamental variables during the whole merging procedure.
\end{enumerate}

The above critical conclusions are expected to benefit the decision-making algorithm design of autonomous vehicles when interacting with human-driven vehicles. For example, the reveal {\textcircled{\small 1}} provides evidence that an autonomous vehicle needs to take the surrounding vehicle's social preference into account to make associated merge decisions. However, representing all environmental features in the real world leads to a combinatorial explosion, yielding too many states, known as the curse of dimensionality. Fortunately, the third finding in Section V-C-3), i.e., `drivers always rely on certain basis variables during the whole merging procedure', guides selecting significant variables in a low dimension to make efficient decisions while not destroying the model performance. Our latest work \cite{wang2021uncovering} succeeded in applying this critical conclusion to building an efficient model, i.e., only using basis variables to improve model performance. The findings {\textcircled{\small 2}} indicate that humans make efficient decisions via a selective attention mechanism, which matches well with laboratory findings on learning mechanisms \cite{gershman2010learning}. In other words, a human-level efficient algorithm for merging behavior at highway on-ramps should select the related variables to make decisions.

\ifCLASSOPTIONcaptionsoff
  \newpage
\fi




\bibliographystyle{IEEEtran.bst}
% argument is your BibTeX string definitions and bibliography 
\bibliography{reference}

\begin{thebibliography}{10}
\providecommand{\url}[1]{#1}
\csname url@samestyle\endcsname
\providecommand{\newblock}{\relax}
\providecommand{\bibinfo}[2]{#2}
\providecommand{\BIBentrySTDinterwordspacing}{\spaceskip=0pt\relax}
\providecommand{\BIBentryALTinterwordstretchfactor}{4}
\providecommand{\BIBentryALTinterwordspacing}{\spaceskip=\fontdimen2\font plus
\BIBentryALTinterwordstretchfactor\fontdimen3\font minus
  \fontdimen4\font\relax}
\providecommand{\BIBforeignlanguage}[2]{{%
\expandafter\ifx\csname l@#1\endcsname\relax
\typeout{** WARNING: IEEEtran.bst: No hyphenation pattern has been}%
\typeout{** loaded for the language `#1'. Using the pattern for}%
\typeout{** the default language instead.}%
\else
\language=\csname l@#1\endcsname
\fi
#2}}
\providecommand{\BIBdecl}{\relax}
\BIBdecl

\bibitem{zgonnikov2020should}
A.~Zgonnikov, D.~Abbink, and G.~Markkula, ``Should i stay or should i go?
  evidence accumulation drives decision making in human drivers,'' 2020.

\bibitem{marczak2013merging}
F.~Marczak, W.~Daamen, and C.~Buisson, ``Merging behaviour: Empirical
  comparison between two sites and new theory development,''
  \emph{Transportation Research Part C: Emerging Technologies}, vol.~36, pp.
  530--546, 2013.

\bibitem{sun2014modeling}
J.~Sun, J.~Ouyang, and J.~Yang, ``Modeling and analysis of merging behavior at
  expressway on-ramp bottlenecks,'' \emph{Transportation Research Record}, vol.
  2421, no.~1, pp. 74--81, 2014.

\bibitem{national2018summary}
N.~H. T.~S. Administration, ``Summary of motor vehicle crashes: 2016 data,''
  United States. National Highway Traffic Safety Administration, Tech. Rep.,
  2018.

\bibitem{endsley1995measurement}
M.~R. Endsley, ``Measurement of situation awareness in dynamic systems,''
  \emph{Human factors}, vol.~37, no.~1, pp. 65--84, 1995.

\bibitem{schwarting2019social}
W.~Schwarting, A.~Pierson, J.~Alonso-Mora, S.~Karaman, and D.~Rus, ``Social
  behavior for autonomous vehicles,'' \emph{Proceedings of the National Academy
  of Sciences}, vol. 116, no.~50, pp. 24\,972--24\,978, 2019.

\bibitem{koechlin2020human}
E.~Koechlin, ``Human decision-making beyond the rational decision theory,''
  \emph{Trends in Cognitive Sciences}, vol.~24, no.~1, pp. 4--6, 2020.

\bibitem{seo2012neural}
H.~Seo and D.~Lee, ``Neural basis of learning and preference during social
  decision-making,'' \emph{Current opinion in neurobiology}, vol.~22, no.~6,
  pp. 990--995, 2012.

\bibitem{leong2017dynamic}
Y.~C. Leong, A.~Radulescu, R.~Daniel, V.~DeWoskin, and Y.~Niv, ``Dynamic
  interaction between reinforcement learning and attention in multidimensional
  environments,'' \emph{Neuron}, vol.~93, no.~2, pp. 451--463, 2017.

\bibitem{niv2019learning}
Y.~Niv, ``Learning task-state representations,'' \emph{Nature neuroscience},
  vol.~22, no.~10, pp. 1544--1553, 2019.

\bibitem{radulescu2021human}
A.~Radulescu, Y.~S. Shin, and Y.~Niv, ``Human representation learning,''
  \emph{Annual Review of Neuroscience}, vol.~44, 2021.

\bibitem{weng2018time}
J.~Weng, G.~Du, D.~Li, and Y.~Yu, ``Time-varying mixed logit model for vehicle
  merging behavior in work zone merging areas,'' \emph{Accident Analysis \&
  Prevention}, vol. 117, pp. 328--339, 2018.

\bibitem{langdon2019uncovering}
A.~J. Langdon, M.~Song, and Y.~Niv, ``Uncovering the ‘state’: Tracing the
  hidden state representations that structure learning and decision-making,''
  \emph{Behavioural processes}, vol. 167, p. 103891, 2019.

\bibitem{yang2018scene}
S.~Yang, W.~Wang, C.~Liu, and W.~Deng, ``Scene understanding in deep
  learning-based end-to-end controllers for autonomous vehicles,'' \emph{IEEE
  Transactions on Systems, Man, and Cybernetics: Systems}, vol.~49, no.~1, pp.
  53--63, 2018.

\bibitem{wilson2012inferring}
R.~C. Wilson and Y.~Niv, ``Inferring relevance in a changing world,''
  \emph{Frontiers in human neuroscience}, vol.~5, p. 189, 2012.

\bibitem{ahmed1996models}
K.~Ahmed, M.~Ben-Akiva, H.~Koutsopoulos, and R.~Mishalani, ``Models of freeway
  lane changing and gap acceptance behavior,'' \emph{Transportation and traffic
  theory}, vol.~13, pp. 501--515, 1996.

\bibitem{lee2006modeling}
G.~Lee, ``Modeling gap acceptance at freeway merges,'' Ph.D. dissertation,
  Massachusetts Institute of Technology, 2006.

\bibitem{toledo2009estimation}
T.~Toledo, H.~N. Koutsopoulos, and M.~Ben-Akiva, ``Estimation of an integrated
  driving behavior model,'' \emph{Transportation Research Part C: Emerging
  Technologies}, vol.~17, no.~4, pp. 365--380, 2009.

\bibitem{sun2014driver}
D.~Sun and L.~Elefteriadou, ``A driver behavior-based lane-changing model for
  urban arterial streets,'' \emph{Transportation science}, vol.~48, no.~2, pp.
  184--205, 2014.

\bibitem{zheng2014recent}
Z.~Zheng, ``Recent developments and research needs in modeling lane changing,''
  \emph{Transportation research part B: methodological}, vol.~60, pp. 16--32,
  2014.

\bibitem{tang2018deviation}
W.~Tang and D.~M. Levinson, ``Deviation between actual and shortest travel time
  paths for commuters,'' \emph{Journal of Transportation Engineering, Part A:
  Systems}, vol. 144, no.~8, p. 04018042, 2018.

\bibitem{kita1993effects}
H.~Kita, ``Effects of merging lane length on the merging behavior at expressway
  on-ramps,'' \emph{Transportation and Traffic Theory}, pp. 37--51, 1993.

\bibitem{weng2011modeling}
J.~Weng and Q.~Meng, ``Modeling speed-flow relationship and merging behavior in
  work zone merging areas,'' \emph{Transportation research part C: emerging
  technologies}, vol.~19, no.~6, pp. 985--996, 2011.

\bibitem{fatema2013probabilistic}
T.~Fatema and Y.~Hassan, ``Probabilistic design of freeway entrance
  speed-change lanes considering acceleration and gap acceptance behavior,''
  \emph{Transportation research record}, vol. 2348, no.~1, pp. 30--37, 2013.

\bibitem{weng2015depth}
J.~Weng, S.~Xue, Y.~Yang, X.~Yan, and X.~Qu, ``In-depth analysis of drivers’
  merging behavior and rear-end crash risks in work zone merging areas,''
  \emph{Accident Analysis \& Prevention}, vol.~77, pp. 51--61, 2015.

\bibitem{li2018application}
G.~Li, ``Application of finite mixture of logistic regression for heterogeneous
  merging behavior analysis,'' \emph{Journal of Advanced Transportation}, vol.
  2018, 2018.

\bibitem{mccall2007lane}
J.~C. McCall, D.~P. Wipf, M.~M. Trivedi, and B.~D. Rao, ``Lane change intent
  analysis using robust operators and sparse bayesian learning,'' \emph{IEEE
  Transactions on Intelligent Transportation Systems}, vol.~8, no.~3, pp.
  431--440, 2007.

\bibitem{meng2011improved}
Q.~Meng and J.~Weng, ``An improved cellular automata model for heterogeneous
  work zone traffic,'' \emph{Transportation research part C: emerging
  technologies}, vol.~19, no.~6, pp. 1263--1275, 2011.

\bibitem{arbis2019game}
D.~Arbis and V.~V. Dixit, ``Game theoretic model for lane changing:
  Incorporating conflict risks,'' \emph{Accident Analysis \& Prevention}, vol.
  125, pp. 158--164, 2019.

\bibitem{meng2012classification}
Q.~Meng and J.~Weng, ``Classification and regression tree approach for
  predicting drivers’ merging behavior in short-term work zone merging
  areas,'' \emph{Journal of Transportation Engineering}, vol. 138, no.~8, pp.
  1062--1070, 2012.

\bibitem{hou2013modeling}
Y.~Hou, P.~Edara, and C.~Sun, ``Modeling mandatory lane changing using bayes
  classifier and decision trees,'' \emph{IEEE Transactions on Intelligent
  Transportation Systems}, vol.~15, no.~2, pp. 647--655, 2013.

\bibitem{tang2018lane}
J.~Tang, F.~Liu, W.~Zhang, R.~Ke, and Y.~Zou, ``Lane-changes prediction based
  on adaptive fuzzy neural network,'' \emph{Expert Systems with Applications},
  vol.~91, pp. 452--463, 2018.

\bibitem{broz2008planning}
F.~Broz \emph{et~al.}, ``Planning for human-robot interaction: representing
  time and human intention,'' Ph.D. dissertation, Carnegie Mellon University,
  The Robotics Institute, 2008.

\bibitem{kirby2010social}
R.~Kirby, \emph{Social robot navigation}.\hskip 1em plus 0.5em minus
  0.4em\relax Carnegie Mellon University, 2010.

\bibitem{wei2013autonomous}
J.~Wei, J.~M. Dolan, and B.~Litkouhi, ``Autonomous vehicle social behavior for
  highway entrance ramp management,'' in \emph{2013 IEEE Intelligent Vehicles
  Symposium (IV)}.\hskip 1em plus 0.5em minus 0.4em\relax IEEE, 2013, pp.
  201--207.

\bibitem{sun2018courteous}
L.~Sun, W.~Zhan, M.~Tomizuka, and A.~D. Dragan, ``Courteous autonomous cars,''
  in \emph{2018 IEEE/RSJ International Conference on Intelligent Robots and
  Systems (IROS)}.\hskip 1em plus 0.5em minus 0.4em\relax IEEE, 2018, pp.
  663--670.

\bibitem{ren2019shall}
Y.~Ren, S.~Elliott, Y.~Wang, Y.~Yang, and W.~Zhang, ``How shall i drive?
  interaction modeling and motion planning towards empathetic and
  socially-graceful driving,'' in \emph{2019 International Conference on
  Robotics and Automation (ICRA)}.\hskip 1em plus 0.5em minus 0.4em\relax IEEE,
  2019, pp. 4325--4331.

\bibitem{hu2019generic}
Y.~Hu, L.~Sun, and M.~Tomizuka, ``Generic prediction architecture considering
  both rational and irrational driving behaviors,'' in \emph{2019 IEEE
  Intelligent Transportation Systems Conference (ITSC)}.\hskip 1em plus 0.5em
  minus 0.4em\relax IEEE, 2019, pp. 3539--3546.

\bibitem{speidel2019towards}
O.~Speidel, M.~Graf, T.~Phan-Huu, and K.~Dietmayer, ``Towards courteous
  behavior and trajectory planning for automated driving,'' in \emph{2019 IEEE
  Intelligent Transportation Systems Conference (ITSC)}.\hskip 1em plus 0.5em
  minus 0.4em\relax IEEE, 2019, pp. 3142--3148.

\bibitem{oh2010estimation}
C.~Oh and T.~Kim, ``Estimation of rear-end crash potential using vehicle
  trajectory data,'' \emph{Accident Analysis \& Prevention}, vol.~42, no.~6,
  pp. 1888--1893, 2010.

\bibitem{gettman2003surrogate}
D.~Gettman and L.~Head, ``Surrogate safety measures from traffic simulation
  models,'' \emph{Transportation Research Record}, vol. 1840, no.~1, pp.
  104--115, 2003.

\bibitem{cunto2008calibration}
F.~Cunto and F.~F. Saccomanno, ``Calibration and validation of simulated
  vehicle safety performance at signalized intersections,'' \emph{Accident
  analysis \& prevention}, vol.~40, no.~3, pp. 1171--1179, 2008.

\bibitem{weng2014rear}
J.~Weng and Q.~Meng, ``Rear-end crash potential estimation in the work zone
  merging areas,'' \emph{Journal of Advanced Transportation}, vol.~48, no.~3,
  pp. 238--249, 2014.

\bibitem{daamen2010empirical}
W.~Daamen, M.~Loot, and S.~P. Hoogendoorn, ``Empirical analysis of merging
  behavior at freeway on-ramp,'' \emph{Transportation Research Record}, vol.
  2188, no.~1, pp. 108--118, 2010.

\bibitem{wang2009study}
J.-q. Wang, R.-j. Chi, L.~Zhang, K.-q. Li, and T.~Yu, ``Study on forward
  collision warning-avoidance algorithm based on driver characteristics
  adaptation,'' \emph{Journal of Highway and Transportation Research and
  Development}, vol.~26, no. supplement 1, 2009.

\bibitem{weng2015modeling}
J.~Weng, S.~Xue, and X.~Yan, ``Modeling vehicle merging behavior in work zone
  merging areas during the merging implementation period,'' \emph{IEEE
  Transactions on Intelligent Transportation Systems}, vol.~17, no.~4, pp.
  917--925, 2015.

\bibitem{alexiadis2004next}
V.~Alexiadis, J.~Colyar, J.~Halkias, R.~Hranac, and G.~McHale, ``The next
  generation simulation program,'' \emph{Institute of Transportation Engineers.
  ITE Journal}, vol.~74, no.~8, p.~22, 2004.

\bibitem{ramanishka2018toward}
V.~Ramanishka, Y.-T. Chen, T.~Misu, and K.~Saenko, ``Toward driving scene
  understanding: A dataset for learning driver behavior and causal reasoning,''
  in \emph{Proceedings of the IEEE Conference on Computer Vision and Pattern
  Recognition}, 2018, pp. 7699--7707.

\bibitem{chang2019argoverse}
M.-F. Chang, J.~Lambert, P.~Sangkloy, J.~Singh, S.~Bak, A.~Hartnett, D.~Wang,
  P.~Carr, S.~Lucey, D.~Ramanan \emph{et~al.}, ``Argoverse: 3d tracking and
  forecasting with rich maps,'' in \emph{Proceedings of the IEEE Conference on
  Computer Vision and Pattern Recognition}, 2019, pp. 8748--8757.

\bibitem{krajewski2018highd}
R.~Krajewski, J.~Bock, L.~Kloeker, and L.~Eckstein, ``The highd dataset: A
  drone dataset of naturalistic vehicle trajectories on german highways for
  validation of highly automated driving systems,'' in \emph{2018 21st
  International Conference on Intelligent Transportation Systems (ITSC)}.\hskip
  1em plus 0.5em minus 0.4em\relax IEEE, 2018, pp. 2118--2125.

\bibitem{zhan2019interaction}
W.~Zhan, L.~Sun, D.~Wang, H.~Shi, A.~Clausse, M.~Naumann, J.~Kummerle,
  H.~Konigshof, C.~Stiller, A.~de~La~Fortelle \emph{et~al.}, ``Interaction
  dataset: An international, adversarial and cooperative motion dataset in
  interactive driving scenarios with semantic maps,'' \emph{arXiv preprint
  arXiv:1910.03088}, 2019.

\bibitem{bray1985multivariate}
J.~H. Bray, S.~E. Maxwell, and S.~E. Maxwell, \emph{Multivariate analysis of
  variance}.\hskip 1em plus 0.5em minus 0.4em\relax Sage, 1985, no.~54.

\bibitem{song2020learning}
M.~Song, Y.~Niv, and M.~B. Cai, ``Learning what is relevant for rewards via
  value-based serial hypothesis testing,'' in \emph{42nd Annual Meeting of the
  Cognitive Science Society, July}, vol.~29, 2020.

\bibitem{nishi2019merging}
T.~Nishi, P.~Doshi, and D.~Prokhorov, ``Merging in congested freeway traffic
  using multipolicy decision making and passive actor-critic learning,''
  \emph{IEEE Transactions on Intelligent Vehicles}, vol.~4, no.~2, pp.
  287--297, 2019.

\bibitem{bouton2019cooperation}
M.~Bouton, A.~Nakhaei, K.~Fujimura, and M.~J. Kochenderfer, ``Cooperation-aware
  reinforcement learning for merging in dense traffic,'' in \emph{2019 IEEE
  Intelligent Transportation Systems Conference (ITSC)}.\hskip 1em plus 0.5em
  minus 0.4em\relax IEEE, 2019, pp. 3441--3447.

\bibitem{wang2021uncovering}
H.~Wang, W.~Wang, S.~Yuan, and X.~Li, ``Uncovering interpretable internal
  states of merging tasks at highway on-ramps for autonomous driving
  decision-making,'' \emph{arXiv preprint arXiv:2102.07530}, 2021.

\bibitem{karimi2020cooperative}
M.~Karimi, C.~Roncoli, C.~Alecsandru, and M.~Papageorgiou, ``Cooperative
  merging control via trajectory optimization in mixed vehicular traffic,''
  \emph{Transportation Research Part C: Emerging Technologies}, vol. 116, p.
  102663, 2020.

\bibitem{guo2020merging}
J.~Guo, S.~Cheng, and Y.~Liu, ``Merging and diverging impact on mixed traffic
  of regular and autonomous vehicles,'' \emph{IEEE Transactions on Intelligent
  Transportation Systems}, 2020.

\bibitem{moridpour2010modeling}
S.~Moridpour, M.~Sarvi, and G.~Rose, ``Modeling the lane-changing execution of
  multiclass vehicles under heavy traffic conditions,'' \emph{Transportation
  research record}, vol. 2161, no.~1, pp. 11--19, 2010.

\bibitem{gershman2010learning}
S.~Gershman, J.~Cohen, and Y.~Niv, ``Learning to selectively attend,'' in
  \emph{Proceedings of the Annual Meeting of the Cognitive Science Society},
  vol.~32, no.~32, 2010.

\end{thebibliography}



\begin{IEEEbiography}[{\includegraphics[width=1in,height=1.25in,clip,keepaspectratio]{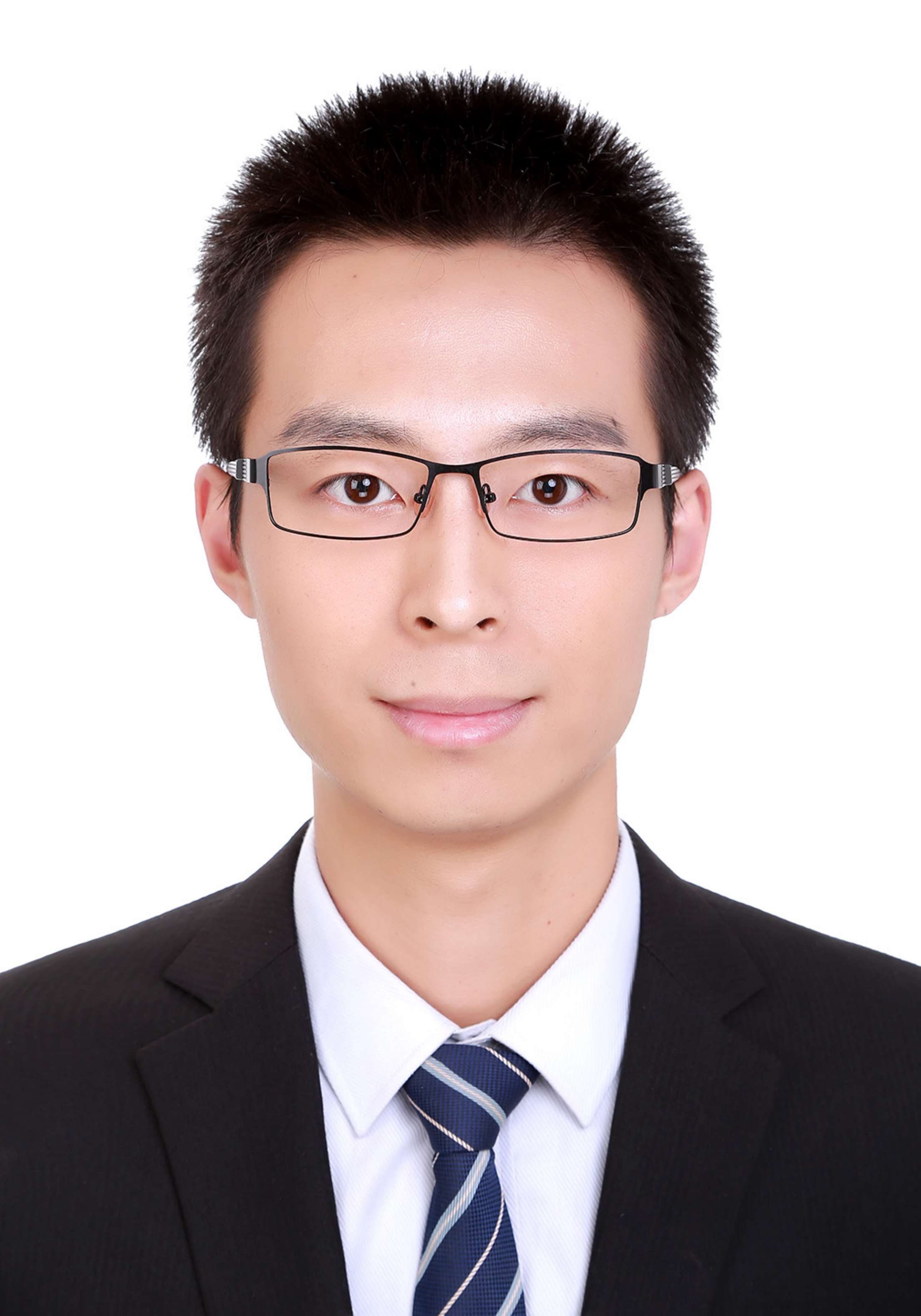}}]{Huanjie Wang}
received the M.S. degree from the School of Mechanical Engineering, Beijing Institute of Technology, China, in 2016, where he is currently pursuing the Ph.D. degree in mechanical engineering. He was also a Research Scholar with the University of California at Berkeley from 2018 to 2020. His research interests include automated vehicles, situational awareness, driver behavior, decision-making, and machine learning. 
\end{IEEEbiography}

\enlargethispage{-1in}

\newpage

\begin{IEEEbiography}[{\includegraphics[width=1in,height=1.25in,clip,keepaspectratio]{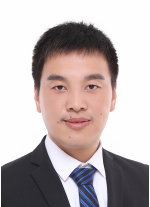}}]{Wenshuo Wang} (SM'15-M'18)
received the Ph.D. degree in mechanical engineering from the Beijing Institute of Technology, Beijing, China, in 2018. He is currently working as a Postdoctoral Researcher with California Partners for Advanced Transportation Technology (California PATH), UC Berkeley. He was a Postdoctoral Research Associate with the Carnegie Mellon University, Pittsburgh, PA, USA, from 2018 to 2019. He was also a Research Scholar with the University of California at Berkeley from 2015 to 2017 and with the University of Michigan, Ann Arbor, from 2017 to 2018. His research interests include Bayesian nonparametric learning, human driver model, human–vehicle interaction, ADAS, and autonomous vehicles.
\end{IEEEbiography}

\begin{IEEEbiography}[{\includegraphics[width=1in,height=1.25in,clip,keepaspectratio]{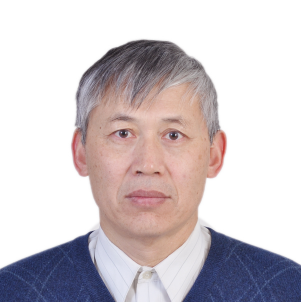}}]{Shihua Yuan}
received the B.S., M.S., and Ph.D. degrees in vehicle engineering from the Beijing Institute of Technology, Beijing, China, in 1982, 1985, and 2000, respectively. From 1992 to 1997, he was an Associate Professor with the Beijing Institute of Technology, where he has been a Professor with the School of Mechanical Engineering, since 1997. He is the author of more than 100 research articles. His research interests include vehicle dynamics, vehicle braking energy recovery, vehicle continuous transmission and its control technology, and unmanned ground vehicle.
\end{IEEEbiography}

\begin{IEEEbiography}[{\includegraphics[width=1in,height=1.25in,clip,keepaspectratio]{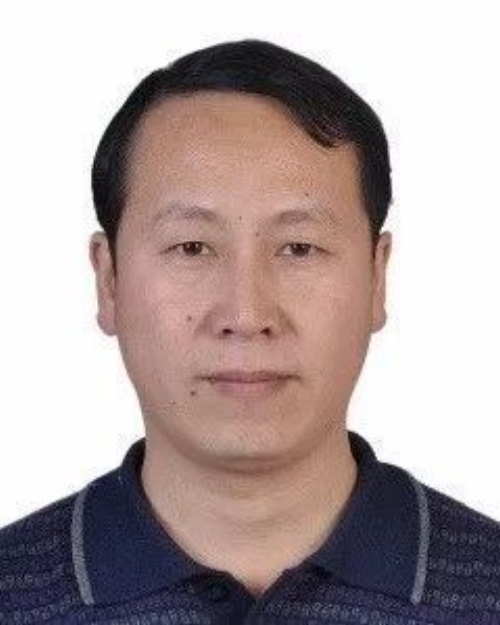}}]{Xueyuan Li}
received the B.S., M.S., and Ph.D. degrees in vehicle engineering from the Beijing Institute of Technology, Beijing, China, in 1999, 2002, and 2010, respectively. He was the Director of the National Key Laboratory of Vehicular Transmission, from 2008 to 2014. He is currently the Vice Director of the Department of Vehicle Engineering, Beijing Institute of Technology. Since 2002, he has been an Associate Professor with the School of Mechanical Engineering, Beijing Institute of Technology. His research interests include vehicle transmission theory and technology, unmanned vehicle theory and technology, and machine learning.
\end{IEEEbiography}

\begin{IEEEbiography}[{\includegraphics[width=1in,height=1.25in,clip,keepaspectratio]{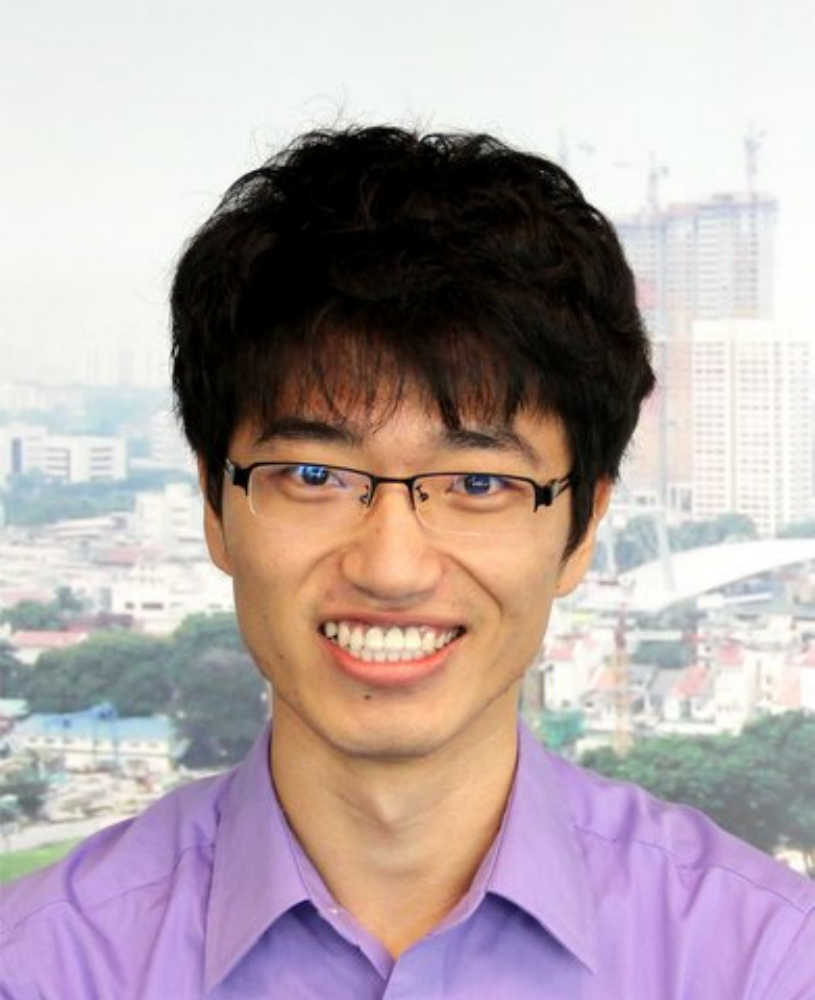}}]{Lijun Sun}
received the B.S. degree in Civil Engineering from Tsinghua University, Beijing, China, in 2011, and Ph.D. degree in Civil Engineering (Transportation) from National University of Singapore in 2015. He is currently an Assistant Professor with the Department of Civil Engineering at McGill University, Montreal, QC, Canada. His research centers on intelligent transportation systems, machine learning, spatiotemporal modeling, travel behavior, and agent-based simulation.
\end{IEEEbiography}

\vfill

\enlargethispage{-1in}

%\newpage

% \begin{IEEEbiographynophoto}{XXX}


%\vfill


\end{document}